\documentclass[10pt,twocolumn,letterpaper]{article}

\usepackage{cvpr}              

\usepackage[dvipsnames]{xcolor}

\usepackage{graphicx}
\usepackage{amsmath}
\usepackage{amssymb}
\usepackage{booktabs}

\usepackage{times}
\usepackage{epsfig}
\usepackage{bbm}

\usepackage{verbatim}
\usepackage{listings}

\usepackage{algorithm}
\usepackage{algorithmicx}
\usepackage{algpseudocode}

\usepackage{color}

\usepackage{ctable}
\usepackage{makecell}
\usepackage{tabularx}
\usepackage{multirow}
\usepackage{multicol}
\usepackage{arydshln}

\usepackage{pifont}
\newcommand{\cmark}{\ding{51}}%
\newcommand{\xmark}{\ding{55}}%

\newcommand{\ourmethod}{HORIA}

\definecolor{cvprblue}{rgb}{0.21,0.49,0.74}
\usepackage[pagebackref,breaklinks,colorlinks,allcolors=cvprblue]{hyperref}

\def\paperID{*****}
\def\confName{CVPR}
\def\confYear{2027}

\title{\ourmethod: 4D Human and Object Reconstruction from a Monocular Video \\ with Interaction-Aware Re-Rendering}

\author{
  Hyeongjin Nam$^{1}$ \hskip1.6em Taeryung Lee$^{2}$ \hskip1.6em Daniel Sungho Jung$^{2}$ \hskip1.6em Kyoung Mu Lee$^{1,2}$ \vspace{+2mm} \\ 
   $^{1}$Dept. of ECE\&ASRI, $^{2}$IPAI, Seoul National University  \\
   {\tt\small \{namhjsnu28, trlee94, dqj5182, kyoungmu\}@snu.ac.kr} \\
   \small{\url{https://hygenie1228.github.io/HORIA/}}
}

\begin{document}
\twocolumn[{
\maketitle
{\centering
\vspace{-0.5em}
\includegraphics[width=0.95\linewidth]{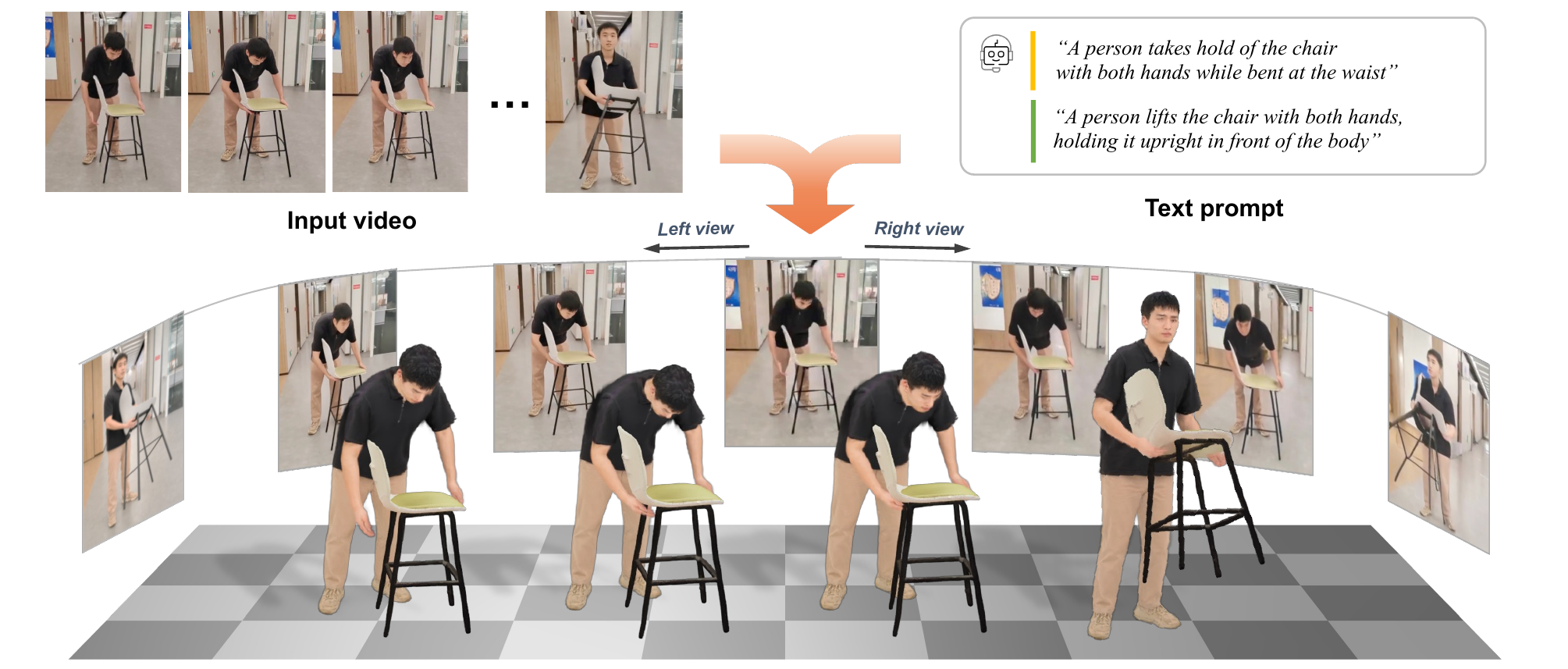}}
\vspace{-0.4em}
\captionof{figure}
{
    \textbf{\ourmethod.}
    Given a monocular video, our framework reconstructs the 4D human and object with spatially and temporally consistent appearance, guided by the text descriptions of each interaction phase.
}
\label{fig:0_teaser}
\vspace{+2.5em}
}]

\begin{abstract}
Joint 4D reconstruction of a human and object from a monocular video is essential for understanding how people interact with their environment.
While recent methods have made noticeable progress, capturing such human-object interactions remains difficult due to two main challenges.
First, a video typically contains composite semantic context spanning multiple interaction phases, making it difficult to track the temporal changes of the interaction in the reconstruction.
Second, under human-object occlusion, a single-view observation contains incomplete visual information about the interaction, resulting in geometrically inconsistent reconstructions.
To address these challenges, we introduce \textit{\textbf{\ourmethod}}, built upon two core designs.
First, we decompose the entire input video into temporally grounded and semantically distinct interaction phases with motion-aware captioning, obtaining a specific text description for each phase.
Second, conditioned on these text descriptions of the interaction, we propose an interaction-aware re-rendering that compensates for the limited single-view observation.
Specifically, we render the initially reconstructed 4D human and object from novel viewpoints and refine the renderings with a text-to-video diffusion prior, which provides auxiliary novel-view supervision beyond single-view cues.
As a result, our framework produces temporally coherent and accurate 4D reconstructions under diverse interaction scenarios, achieving state-of-the-art performance.
\end{abstract}

\section{Introduction}
\label{sec:introduction}
Joint 4D reconstruction of a human and object from a monocular video has garnered significant attention in various applications such as robot manipulation, immersive AR/VR experiences, and digital content creation.
The human and object interact through not only physical relationships, such as contact between their surfaces, but also semantic relationships that characterize the functional intent of the interaction, such as reaching for a cup.
Capturing both relationships of human-object interaction (HOI) is crucial for faithful 4D reconstruction.

While recent approaches~\cite{xu2021d3d,xie2023visibility,xue2026rhino,xie2026cari4d,wen2025efficient} have shown notable progress, they still face two main challenges.
First, existing methods struggle to capture the evolving semantic context of interactions over time.
They rely primarily on local geometric proximity without considering the semantic context, which text descriptions~\cite{nam2026tehor} can incorporate into the reconstruction.
However, a video typically spans multiple interaction phases, and summarizing the entire sequence into a single global description fails to capture which interaction applies at each moment.
Second, a monocular observation misses the geometric cues of the occluded regions, where the human and object frequently obscure each other.
This lack of geometric cues makes their spatial relationship difficult to reconstruct, often leading to severe interpenetration and temporally inconsistent results.

To address the above challenges, we present \textbf{\ourmethod}, a framework that reconstructs the 4D human and object from a monocular video through interaction-aware re-rendering.
Our framework first splits the input video into several interaction phases and assigns a text description to each phase.
This decomposition allows each description to capture the fine-grained temporal dynamics of its own phase.
To decide where to decompose, we estimate the object motion and incorporate it into the captioning of human-object interactions, a process we refer to as motion-aware captioning.
Based on each caption, we split the video into temporally grounded interaction phases with precise boundaries.
Consequently, each text description is aligned with its corresponding interaction phase, providing reliable temporal guidance for accurate 4D reconstruction.

\begin{figure}[t!]
  \centering
  \includegraphics[width=1.0\linewidth]{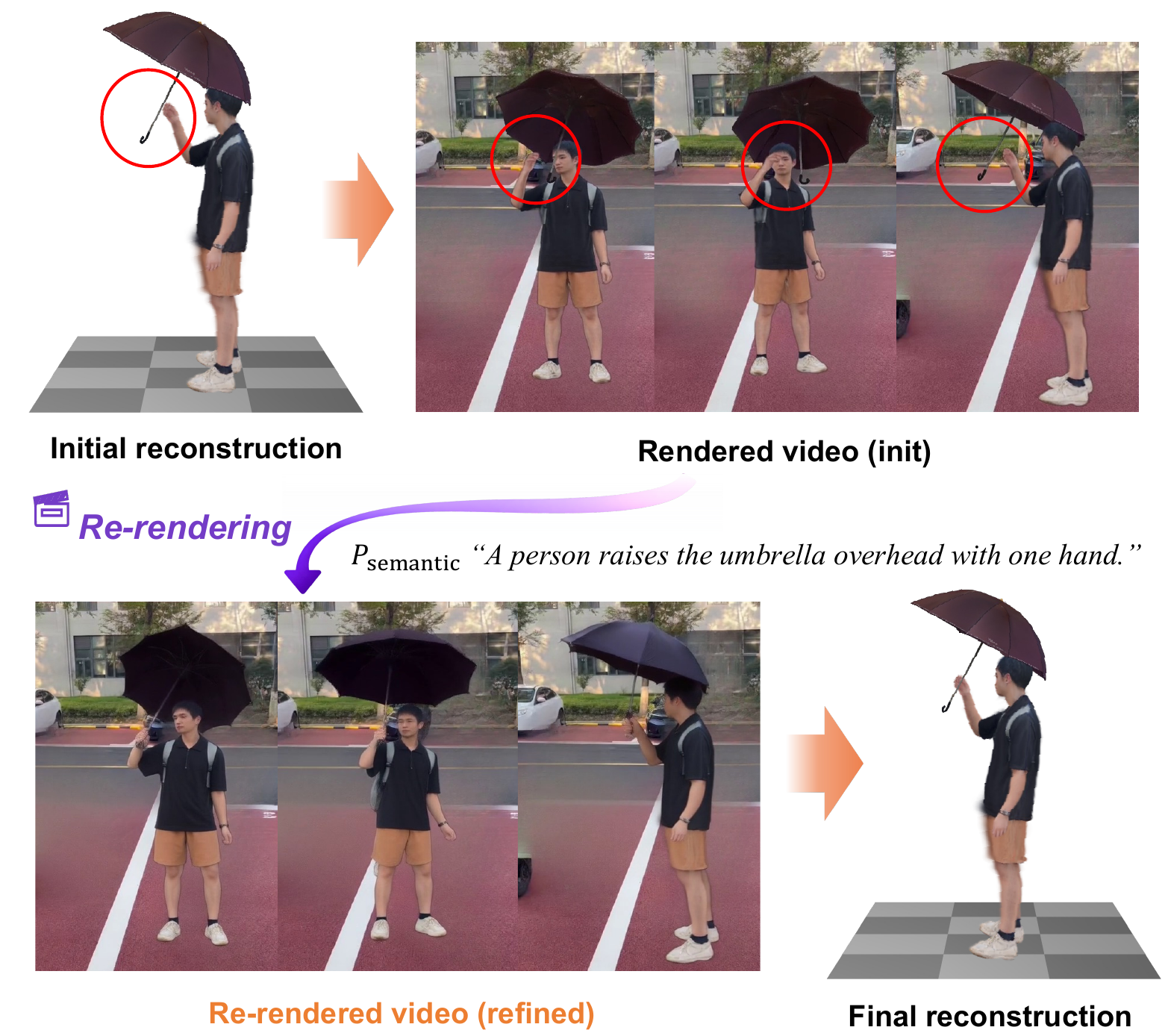}
  \vspace*{-1.8em}
  \caption{\textbf{Re-rendering for reconstruction.}
  Conditioned on the text description $P_{\text{semantic}}$, our re-rendering refines the novel-view renderings of the initial reconstruction, and the refined videos serve as auxiliary supervision for the final 4D reconstruction.
  }
  \vspace*{+0.0em}
  \label{fig:1_introduction}
\end{figure}

Building upon these grounded descriptions, we introduce an interaction-aware re-rendering pipeline that renders and refines novel-view videos of the reconstruction to compensate for limited single-view observations.
Specifically, we render the initially reconstructed human and object from novel viewpoints and feed the renderings into a text-to-video diffusion network~\cite{wan2025wan}.
Leveraging its prior knowledge of human-object interactions, the network refines them toward plausible interaction videos under the text description.
During this refinement, we additionally apply distinct text prompts to the human and object regions, keeping the appearance of each region faithful to its own description and specifying where they interact.
After the re-rendering process, we aggregate the refined novel-view videos into a single signed distance field (SDF), which serves as a geometric consensus across viewpoints.
As shown in \cref{fig:1_introduction}, our re-rendering process corrects the spatial misalignment between the hand and the umbrella, guiding the final reconstruction.
As a result, our framework produces accurate and plausible 4D reconstructions of the human and object through re-rendering that compensates for the limited single-view observation.

Extensive experiments demonstrate that our framework significantly outperforms existing reconstruction methods across various interaction scenarios.
Our contributions can be summarized as follows.
\begin{itemize}
\item We propose \textbf{\ourmethod}, a framework for joint 4D reconstruction of a human and object from a monocular video via temporally grounded novel-view video refinement.
\item To capture fine-grained human-object interactions across the entire video sequence, we decompose the input video into several interaction phases with our proposed motion-aware captioning.
\item To compensate for limited single-view cues from the input video, we introduce an interaction-aware re-rendering pipeline that provides auxiliary novel-view supervision for accurate 4D human and object reconstruction.
\end{itemize}

\section{Related works}
\label{sec:related_works}
\noindent\textbf{Human and object reconstruction.}
Most recent works on human and object reconstruction~\cite{chen2019holistic++,zhang2020perceiving,xie2022chore,nam2024contho,xie2023templatefree,dwivedi2025interactvlm,wen2025reconstructing,li2025scorehoi,xu2021d3d,xie2023visibility,xue2026rhino,xie2025intertrack,xie2026cari4d,wen2025efficient,cseke2025pico,huang2026recovering} reconstruct the 3D human and object from a single image or monocular video, relying on geometric constraints, such as contact maps.
CONTHO~\cite{nam2024contho} jointly predicts dense contact maps from images and refines the reconstruction using a contact-based Transformer~\cite{vaswani2017attention}.
VisTracker~\cite{xie2023visibility} exploits the correlation between human and object motion, predicting the object pose of occluded frames from the human motion and the neighboring visible frames.
RHINO~\cite{xue2026rhino} jointly recovers the human, object, and scene from a monocular video, and refines their poses with contact-driven optimization.
CARI4D~\cite{xie2026cari4d} proposes a render-and-compare network that aligns rendered estimates with input observations to enhance reconstructions.

Unlike existing 4D human and object reconstruction methods, our framework has two key differences.
First, prior works rely primarily on local geometric priors, such as contact maps, which cannot represent the semantic context of the interaction.
Our framework uses text descriptions that carry the semantic context, assigning a description to each interaction phase through video decomposition.
This enables the reconstruction to reflect the complex interactions of the real world.
Second, whereas previous methods reconstruct the human and object from a single viewpoint that suffers from human-object occlusion, our framework refines the novel-view renderings of the reconstruction via interaction-aware re-rendering.
This re-rendering process supplements the occluded regions and serves as auxiliary supervision for accurate reconstruction.

\noindent\textbf{Video decomposition.}
Video decomposition methods localize the segment of an input video that semantically corresponds to a text query~\cite{gao2017tall,zhang2020learning,lei2021moment,moon2023query,lin2023univtg,ren2024timechat,huang2024vtimellm,wang2024groundedvideollm,chen2025fmimic,zhang2026timelens}.
TALL~\cite{gao2017tall} introduces a cross-modal temporal regression method to predict temporal action boundaries aligned with a text query.
TimeLens~\cite{zhang2026timelens} adapts pretrained multimodal large language models~\cite{bai2025qwen3} with interleaved timestamp encoding and reinforcement learning for precise temporal grounding.
FMimic~\cite{chen2025fmimic} decomposes a video into human-object interaction states using contact distance thresholds for robotic imitation learning.
Different from FMimic, our video decomposition takes a coarse-to-fine strategy that detects initial segment boundaries from object trajectories and refines them into precise interaction phase boundaries using TimeLens~\cite{zhang2026timelens}.
This aligns each text description with the interval it describes, providing accurate boundaries of interaction phases for 4D human and object reconstruction.

\noindent\textbf{Generative priors for reconstruction.}
Recently, various reconstruction methods~\cite{deng2023nerdi,huang2024tech,tang2023make,ye2024ghop,tang2024dreamgaussian,sun2024dreamcraft3d,qian2024magic123,wang2024animatabledreamer,nam2026tehor,huang2025vivid4d,liu2023zero} have leveraged generative priors (\textit{e.g.}, text-to-image generation~\cite{rombach2022high}) to supplement the information missing from a single-view observation of the input image or video.
NeRDi~\cite{deng2023nerdi} reconstructs a 3D object from a single image by minimizing a diffusion loss on its novel-view renderings with a text caption.
Vivid4D~\cite{huang2025vivid4d} reconstructs a 4D dynamic scene from a monocular video, warping the video's pixels into novel viewpoints and completing the occlusions with a video inpainting model.
TeHOR~\cite{nam2026tehor} jointly reconstructs a 3D human and object by distilling a diffusion prior conditioned on text descriptions that contain the semantic context of the interaction.
Most of these methods leverage the distillation strategy~\cite{poole2023dreamfusion}, which computes text-conditioned score gradients on the renderings and drives them toward the distribution of the diffusion prior.
However, the score gradients provide an over-smoothed signal~\cite{wang2023prolificdreamer}, which degrades the reconstruction quality.
In contrast, our re-rendering refines the novel-view renderings, which serve as supervision targets, providing clear interaction cues for accurate 4D reconstruction.

\begin{figure*}[t!]
  \centering
  \includegraphics[width=1.0\linewidth]{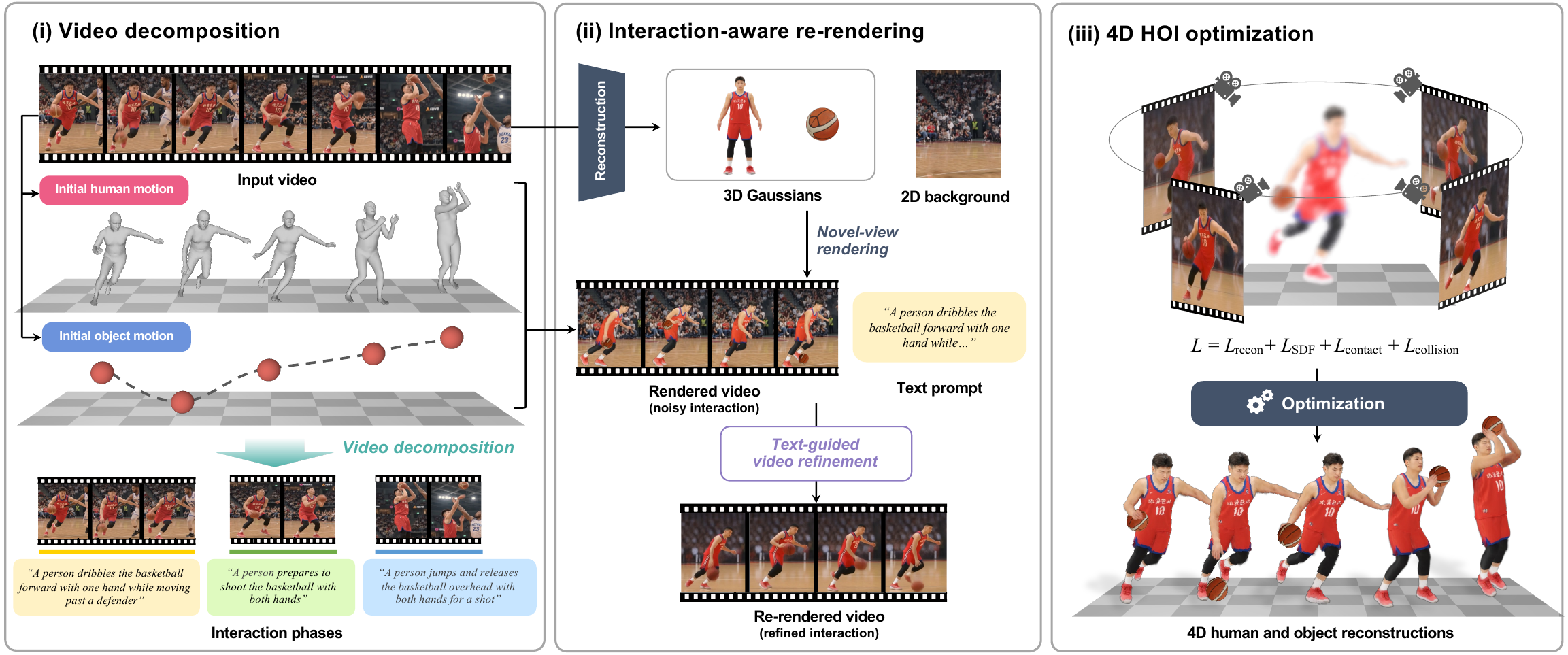}
  \vspace*{-1.3em}
  \caption{\textbf{Overall pipeline of \ourmethod.}
  Given an input video, our framework first decomposes it into interaction phases based on initially estimated human and object motion.
  Then, the text description of each phase guides the re-rendering of the human and object under novel viewpoints and provides auxiliary supervision for 4D HOI optimization.
  }
   \vspace*{+0.0em}
  \label{fig:2_pipeline}
\end{figure*}
\section{\ourmethod}
\label{sec:proposed_method}
\begin{figure*}[t!]
  \centering
  \includegraphics[width=1.0\linewidth]{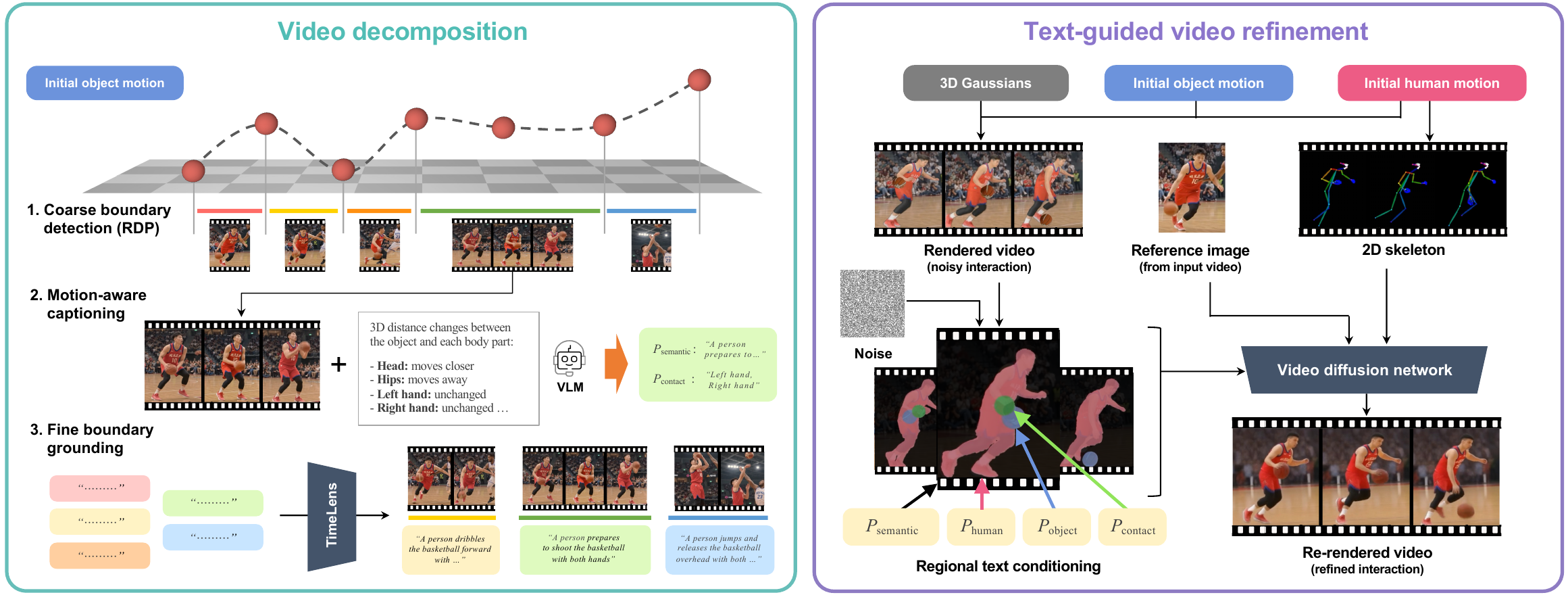}
  \vspace*{-1.3em}
  \caption{\textbf{Detailed process of video decomposition (left) and text-guided video refinement (right).}
  }
   \vspace*{+0.0em}
  \label{fig:3_pipeline_detail}
\end{figure*}

\cref{fig:2_pipeline,fig:3_pipeline_detail} illustrate the overall pipeline of \ourmethod.
We first describe the 4D representation of the human and object (\cref{sec:3.1_representation}).
Subsequently, we describe three stages: video decomposition (\cref{sec:3.2_decomposition}), interaction-aware re-rendering (\cref{sec:3.3_rerendering}), and 4D HOI optimization (\cref{sec:3.4_optimization}).

\subsection{4D representation}
\label{sec:3.1_representation}
Given a monocular video $X$ of $T$ frames, we reconstruct the motion of the human body and the rigid object over the entire sequence.
We represent the human and object as sets of 3D Gaussians~\cite{kerbl3Dgaussians}, denoted by $\{\Phi_{\text{h},t}\}_{t=1}^{T}$ and $\{\Phi_{\text{o},t}\}_{t=1}^{T}$, which can be rendered into 2D space through a differentiable rasterizer~\cite{yu2024mip}.
Each Gaussian is anchored to a surface point of its base mesh, which allows the Gaussians to be converted into meshes~\cite{nam2026tehor}.

\noindent\textbf{Human representation.}
We represent the human with the SMPL-X model~\cite{pavlakos2019expressive}, parameterized by the human pose $\theta$, shape $\beta$, and translation $\gamma$.
We attach a set of 3D Gaussians $\Phi_{\text{h}}$ to the canonical SMPL-X mesh.
These Gaussians are deformed via linear blend skinning with $\theta$~\cite{moon2024expressive}.

\noindent\textbf{Object representation.}
We represent the rigid object with a set of 3D Gaussians $\Phi_{\text{o}}$ defined in a canonical space.
These canonical Gaussians are then transformed into 3D space via a similarity transformation parameterized by rotation $R$, translation $\mathcal{T}$, and scale $s$.

\noindent\textbf{Temporal parameterization.}
We parameterize the motion of both the human and object using a cubic B-spline~\cite{de1978practical}, a polynomial curve determined by a set of coefficients defined at sparse time intervals.
Unlike independent per-frame pose parameters, the B-spline assigns shared coefficients to neighboring frames, yielding smooth motion.
Specifically, we define the concatenated motion sequence of the human and object as $\Psi=[\theta_t,\gamma_t,R_t,\mathcal{T}_t]_{t=1}^{T} \in \mathbb{R}^{T \times D}$ and formulate it as $\Psi = \mathbf{B}\mathbf{C}$.
$\mathbf{B}\in\mathbb{R}^{T\times N}$ is a B-spline basis matrix, and $\mathbf{C}\in\mathbb{R}^{N\times D}$ is a coefficient matrix, where $N$ is the number of coefficients and $D$ is the dimension of the motion parameters.
All rotations use the continuous 6D rotation representation~\cite{zhou2019continuity}.

\subsection{Video decomposition}
\label{sec:3.2_decomposition}
We decompose the input video into interaction phases by identifying boundaries where the interaction changes based on the human and object motions.
Our video decomposition takes a coarse-to-fine strategy: we first detect coarse boundaries from the object trajectory, caption each coarse segment, and then refine the boundaries by grounding each caption in the video.
Before the video decomposition, we independently extract the initial motions of the human and object using Multi-HMR~\cite{baradel2024multi} and FoundationPose~\cite{wen2024foundationpose} respectively.
These initial motions serve as physical cues for tracking changes in the relative distance between the human and object over time.

\noindent\textbf{Coarse boundary detection.}
Based on the initial motion estimations, we first detect coarse boundaries where the object trajectory changes sharply.
To this end, we simplify the object trajectory using the Ramer-Douglas-Peucker (RDP) algorithm~\cite{douglas1973algorithms}.
RDP is a curve simplification method that finds coarse boundaries where the trajectory departs sharply from the overall path.
Such a sharp change in the object trajectory typically coincides with a change in the interaction, such as the object being picked up, carried, or put down.

\noindent\textbf{Motion-aware captioning.}
From each coarse segment, we extract two text prompts, $P_{\text{semantic}}$ and $P_{\text{contact}}$.
$P_{\text{semantic}}$ captures the global context of the interaction phase, providing detailed descriptions of human appearance, object category, and their interaction type.
$P_{\text{contact}}$ specifies the human body parts in physical contact with the object during that interval.
These prompts are acquired using a vision-language model (VLM)~\cite{openai2026gpt5} from five uniformly sampled frames of each coarse segment.
To incorporate 3D human-object distance information in the captioning, we measure the 3D distance between the object and several human parts (\textit{e.g.}, head and hands).
Then, we state whether the object \textit{moves closer}, \textit{moves away}, or stays \textit{unchanged} relative to each part from the start to the end of the segment, and provide this information as additional input to the VLM.
Details are provided in the supplementary material.

\noindent\textbf{Fine boundary grounding.}
Based on the coarse segments and their text descriptions, we refine the coarse boundaries into precise temporal spans using a temporal grounding model, TimeLens~\cite{zhang2026timelens}.
Specifically, we query TimeLens with each caption across the full video timeline to accurately localize the visual extent of each interaction.
If predicted intervals overlap, we merge them into a single continuous segment.
The resulting segment intervals serve as the final interaction phases used for the re-rendering stage.

\subsection{Interaction-aware re-rendering}
\label{sec:3.3_rerendering}
For each interaction phase, we render the 3D human and object Gaussians from fixed novel viewpoints and refine the renderings with a text-to-video diffusion network~\cite{wan2025wan}.
To this end, we reconstruct the canonical human Gaussians $\Phi_{\text{h}}$ using LHM~\cite{qiu2025lhm} and the base object Gaussians $\Phi_{\text{o}}$ using SAM 3D~\cite{chen2026sam}, respectively.
We also build a 2D background by removing the human and object from the first frame using SmartEraser~\cite{jiang2025smarteraser}.

\noindent\textbf{Novel-view rendering and refinement.}
For each novel-view camera $\pi$, sampled at $30^\circ$ azimuth intervals, we render the RGB appearance of the posed human and object Gaussians onto the 2D background to obtain a novel-view video $X_{\pi}$, and project the SMPL-X joints from the same viewpoint to obtain a skeleton video $S_{\pi}$.
We encode $X_{\pi}$ and inject noise to an initial noise level $\tau_0$ to obtain $X_{\tau_0,\pi}$, where the sampling process starts from the rendered appearance.
Conditioned on the semantic prompt $P_{\text{semantic}}$ and a reference image $X_{\text{ref}}$, which is the first frame of the interaction phase, we refine $X_{\tau_0,\pi}$ into
\begin{equation}
X^{*}_{\pi} = f_\psi\big(X_{\tau_0,\pi}; P_{\text{semantic}}, S_{\pi}, X_{\text{ref}}\big)
\label{eq:refine}
\end{equation}
where $f_\psi$ denotes a stochastic sampling process~\cite{lipman2023flow} built on a pretrained video diffusion network~\cite{wan2025wan}.
Using the prior knowledge of the diffusion network learned from large-scale video data, the stochastic sampling effectively corrects artifacts associated with implausible human-object interactions.

\noindent\textbf{Regional text conditioning.}
To explicitly guide the human and object regions during the refinement, we condition each region on its own text prompt: $P_{\text{human}}$ for the human mask $M_{\text{human}}$ and $P_{\text{object}}$ for the object mask $M_{\text{object}}$.
The masks are obtained by rendering the human and object Gaussians separately.
$P_{\text{human}}$ and $P_{\text{object}}$ are extracted from the semantic prompt $P_{\text{semantic}}$, and the extraction details are provided in the supplementary material.
To capture the contact region, we draw a circular region around the joints of $S_{\pi}$ specified by $P_{\text{contact}}$ and merge it into the object mask $M_{\text{object}}$.
Within every cross-attention layer of the sampling process in \cref{eq:refine}, we blend each region's prompt into the base prompt $P_{\text{semantic}}$ as follows:
\begin{multline}
\text{Attn}^\prime(X_{\tau,\pi}) = (1-\lambda M_{\text{union}})\,\text{Attn}(X_{\tau,\pi}; P_{\text{semantic}}) \\
+ \lambda \sum\nolimits_{r} M_r\,\text{Attn}(X_{\tau,\pi}; P_r),
\label{eq:regional_text}
\end{multline}
where $r \in \{\text{human}, \text{object}\}$ indexes the two regions, $M_{\text{union}} = M_{\text{human}} \cup M_{\text{object}}$ is the union of their masks, and $\lambda = 0.2$ is a blend weight.
This regional conditioning prevents semantic interference between the human and object while preserving their interaction.

\subsection{4D HOI optimization}
\label{sec:3.4_optimization}
Based on the re-rendered novel-view videos, we jointly optimize the motion sequence $\Psi$ to recover the accurate 4D human and object.
The overall loss formulation is
\begin{equation}
\label{eq:overall_loss}
\mathcal{L}=\mathcal{L}_{\text{recon}}+\mathcal{L}_{\text{SDF}}+\mathcal{L}_{\text{contact}}+\mathcal{L}_{\text{collision}},
\end{equation}
where $\mathcal{L}_\text{collision}$ is a penalization term that discourages interpenetration between the human and object Gaussians, following \cite{jiang2020coherent}.
The remaining terms are detailed below.

\noindent\textbf{Reconstruction loss.}
The reconstruction loss $\mathcal{L}_{\text{recon}}$ measures the discrepancy between the rendered human and object sequence and two types of target videos: the input video $X$ and the re-rendered novel-view videos $X^{*}_{\pi}$.
Specifically, it computes the mean squared error (MSE) on both RGB pixel values and foreground silhouette masks across all camera viewpoints.

\noindent\textbf{SDF loss.}
The signed distance function (SDF) loss $\mathcal{L}_{\text{SDF}}$ enforces geometric consistency by aggregating the novel-view evidence into a single 3D point cloud for each frame.
As the novel-view re-rendering operates on each viewpoint independently, relying solely on $\mathcal{L}_{\text{recon}}$ can result in inconsistent visual cues across views.
To resolve this problem, we construct a unified 3D geometric consensus across all viewpoints via an SDF.
Specifically, we extract object segmentation masks~\cite{ravi2025sam} and metric depth maps~\cite{piccinelli2024unidepth} from each re-rendered video $X^{*}_{\pi}$ and unproject the masked depth maps into 3D point clouds across all views.
We then aggregate the point clouds and remove outliers via DBSCAN clustering~\cite{ester1996density}, suppressing noisy points inconsistent with the majority of views.
From the point cloud, we build a distance field and align the object Gaussians with the field.
We apply this loss to the object Gaussians only.

\begin{table}[t]
\def\arraystretch{1.6}
\renewcommand{\tabcolsep}{1.0mm}
\footnotesize
\begin{center}
\scalebox{0.65}{
    \begin{tabular}{
    >{\centering\arraybackslash}m{2.6cm}
    >{\centering\arraybackslash}m{2.1cm}|
    >{\centering\arraybackslash}m{0.9cm}
    >{\centering\arraybackslash}m{0.9cm}
    >{\centering\arraybackslash}m{1.1cm}
    >{\centering\arraybackslash}m{1.2cm}
    >{\centering\arraybackslash}m{1.1cm}
    >{\centering\arraybackslash}m{1.1cm}
    }
    \specialrule{.1em}{.05em}{0.0em}
    Caption & Decomposition & CD$_{\text{H}}{\downarrow}$ & CD$_{\text{O}}{\downarrow}$ & Contact${\uparrow}$ & Collision${\downarrow}$ & Accel$_{\text{H}}{\downarrow}$ & Accel$_{\text{O}}{\downarrow}$ \\
    \hline
    \xmark      & \xmark    & 5.723 & 54.505 & 0.349 & 0.072 & 0.307 & 3.468 \\
    Standard    & \xmark    & 5.500 & 51.155 & 0.391 & 0.059 & 0.291 & 3.129 \\
    \textbf{Motion-aware}   & \xmark & 5.511 & 42.102 & 0.408 & 0.060 & 0.294 & 3.210 \\
    \textbf{Motion-aware}   & Uniform & 5.404 & 37.121 & 0.469 & \textbf{0.056} & 0.298 & 2.986 \\
    \makebox[0pt][c]{\textbf{\;\;\;\;Motion-aware\;\;(Ours)}} & \makebox[0pt][c]{\textbf{Grounded}} & \textbf{5.385} & \textbf{34.894} & \textbf{0.520} & \textbf{0.056} & \textbf{0.273} & \textbf{1.571} \\
    \specialrule{.1em}{-0.05em}{-0.05em}
    \end{tabular}
    \vspace*{-1mm}
}
\end{center}
    \vspace*{-1.4em}
    \caption{
    \textbf{Effect of video decomposition.}
    }
    \vspace*{-0.0em}
    \label{tab:abl_decomposition}
\end{table}
\begin{table}[t]
\def\arraystretch{1.6}
\renewcommand{\tabcolsep}{1.0mm}
\footnotesize
\begin{center}
\scalebox{0.65}{
    \begin{tabular}{
    >{\centering\arraybackslash}m{2.0cm}
    >{\centering\arraybackslash}m{2.3cm}|
    >{\centering\arraybackslash}m{0.9cm}
    >{\centering\arraybackslash}m{0.9cm}
    >{\centering\arraybackslash}m{1.1cm}
    >{\centering\arraybackslash}m{1.2cm}
    >{\centering\arraybackslash}m{1.1cm}
    >{\centering\arraybackslash}m{1.1cm}
    }
    \specialrule{.1em}{.05em}{0.0em}
    Re-rendering & Regional cond. & CD$_{\text{H}}{\downarrow}$ & CD$_{\text{O}}{\downarrow}$ & Contact${\uparrow}$ & Collision${\downarrow}$ & Accel$_{\text{H}}{\downarrow}$ & Accel$_{\text{O}}{\downarrow}$ \\
    \hline
    \xmark & \xmark & 5.490 & 64.176 & 0.342 & 0.064 & 0.289 & 3.121 \\
    \cmark & \xmark & 5.452 & 35.923 & 0.511 & 0.059 & 0.291 & 1.794 \\
    \textbf{\cmark\makebox[0pt][l]{\;\;\;\;\;(Ours)}} & \textbf{\cmark} & \textbf{5.385} & \textbf{34.894} & \textbf{0.520} & \textbf{0.056} & \textbf{0.273} & \textbf{1.571} \\
    \specialrule{.1em}{-0.05em}{-0.05em}
    \end{tabular}
    \vspace*{-1mm}
}
\end{center}
    \vspace*{-1.4em}
    \caption{
    \textbf{Effect of re-rendering.}
    }
    \vspace*{-0.0em}
    \label{tab:abl_rerendering}
\end{table}
\begin{table}[t]
\def\arraystretch{1.6}
\renewcommand{\tabcolsep}{1.0mm}
\footnotesize
\begin{center}
\scalebox{0.67}{
    \begin{tabular}{
    >{\raggedright\arraybackslash}m{3.2cm}|
    >{\centering\arraybackslash}m{1.1cm}
    >{\centering\arraybackslash}m{1.1cm}
    >{\centering\arraybackslash}m{1.2cm}
    >{\centering\arraybackslash}m{1.4cm}
    >{\centering\arraybackslash}m{1.2cm}
    >{\centering\arraybackslash}m{1.2cm}
    }
    \specialrule{.1em}{.05em}{0.0em}
    Methods & CD$_{\text{H}}{\downarrow}$ & CD$_{\text{O}}{\downarrow}$ & Contact${\uparrow}$ & Collision${\downarrow}$ & Accel$_{\text{H}}{\downarrow}$ & Accel$_{\text{O}}{\downarrow}$ \\
    \hline
    Initial reconstruction & 5.821 & 88.508 & 0.275 & 0.069 & 0.334 & 20.034 \\
    w/o $\mathcal{L}_{\text{SDF}}$ & 5.387 & 50.162 & 0.344 & 0.061 & 0.274 & 2.761 \\
    w/o $\mathcal{L}_{\text{contact}}$ & 5.418 & 40.162 & 0.459 & 0.057 & 0.288 & 1.709 \\
    w/o $\mathcal{L}_{\text{collision}}$ & 5.390 & 38.945 & 0.512 & 0.061 & 0.277 & 1.602 \\
    \textbf{\ourmethod~(Ours)} & \textbf{5.385} & \textbf{34.894} & \textbf{0.520} & \textbf{0.056} & \textbf{0.273} & \textbf{1.571} \\
    \specialrule{.1em}{-0.05em}{-0.05em}
    \end{tabular}
    \vspace*{-1mm}
}
\end{center}
    \vspace*{-1.4em}
    \caption{
    \textbf{Ablation on 4D HOI optimization.}
    }
    \vspace*{-0.0em}
    \label{tab:abl_optimization}
\end{table}

\noindent\textbf{Contact loss.}
The contact loss $\mathcal{L}_{\text{contact}}$ encourages geometric proximity between the object surface and the human body parts specified in $P_{\text{contact}}$.
We identify the Gaussian centers on the SMPL-X body model corresponding to these parts and minimize the distance between them and their nearest object points.
This loss enforces local physical plausibility across the estimated interaction regions.

\section{Experiments}
\label{sec:experiments}
\begin{table*}[t]
\def\arraystretch{1.6}
\renewcommand{\tabcolsep}{0.6mm}
\footnotesize
\begin{center}
\scalebox{0.85}{
    \begin{tabular}{
    >{\centering\arraybackslash}m{0.4cm}
    >{\raggedright\arraybackslash}m{2.65cm}|
    >{\centering\arraybackslash}m{1.2cm}
    >{\centering\arraybackslash}m{1.2cm}
    >{\centering\arraybackslash}m{1.35cm}
    >{\centering\arraybackslash}m{1.5cm}
    >{\centering\arraybackslash}m{1.3cm}
    >{\centering\arraybackslash}m{1.3cm}|
    >{\centering\arraybackslash}m{1.2cm}
    >{\centering\arraybackslash}m{1.2cm}
    >{\centering\arraybackslash}m{1.35cm}
    >{\centering\arraybackslash}m{1.5cm}
    >{\centering\arraybackslash}m{1.3cm}
    >{\centering\arraybackslash}m{1.3cm}
    }
    \specialrule{.1em}{.05em}{0.0em}

        \multicolumn{2}{l|}{}
        & \multicolumn{6}{c|}{Open4DHOI~\cite{wen2025efficient}}
        & \multicolumn{6}{c}{BEHAVE~\cite{bhatnagar2022behave}} \\

        \multicolumn{2}{l|}{Methods}
        & $\text{CD}_{\text{H}}{\downarrow}$
        & $\text{CD}_{\text{O}}{\downarrow}$
        & $\text{Contact}{\uparrow}$
        & $\text{Collision}{\downarrow}$
        & $\text{Accel}_{\text{H}}{\downarrow}$
        & $\text{Accel}_{\text{O}}{\downarrow}$
        & $\text{CD}_{\text{H}}{\downarrow}$
        & $\text{CD}_{\text{O}}{\downarrow}$
        & $\text{Contact}{\uparrow}$
        & $\text{Collision}{\downarrow}$
        & $\text{Accel}_{\text{H}}{\downarrow}$
        & $\text{Accel}_{\text{O}}{\downarrow}$ \\
        \hline

        \multirow{4}{*}{\rotatebox{90}{Image-based\;\;}}
        & PHOSA~\cite{zhang2020perceiving}
        & 6.434 & 61.138 & 0.267 & \underline{0.061} & -- & --
        & 5.661 & 40.355 & 0.326 & 0.099 & -- & -- \\

        & HOI-Gaussian~\cite{wen2025reconstructing}
        & \textbf{5.330} & 47.086 & 0.382 & 0.072 & -- & --
        & \textbf{3.420} & 39.722 & 0.521 & \underline{0.097} & -- & -- \\

        & TeHOR~\cite{nam2026tehor}
        & 5.501 & \underline{44.397} & \underline{0.456} & 0.070 & -- & --
        & 3.627 & \underline{31.168} & \underline{0.552} & 0.099 & -- & -- \\

        & \textbf{\ourmethod~(Ours)}
        & \underline{5.380} & \textbf{37.268} & \textbf{0.501} & \textbf{0.056} & -- & --
        & \underline{3.512} & \textbf{13.013} & \textbf{0.569} & \textbf{0.080} & -- & -- \\

        \hline

        \multirow{4}{*}{\rotatebox{90}{Video-based\;\;}}
        & VisTracker~\cite{xie2023visibility}
        & 6.392 & 51.289 & 0.383 & \underline{0.152} & \underline{0.459} & \underline{2.197}
        & 6.049 & 16.664 & 0.464 & \underline{0.108} & 1.553 & 1.336 \\

        & InterTrack~\cite{xie2025intertrack}
        & 7.143 & 49.110 & 0.482 & 0.218 & 2.534 & 5.996
        & 9.487 & 19.441 & 0.560 & 0.268 & 2.258 & 5.427 \\

        & CARI4D~\cite{xie2026cari4d}
        & \underline{5.740} & \underline{39.090} & \underline{0.514} & 0.182 & 0.710 & 2.332
        & \underline{3.582} & \underline{14.213} & \underline{0.583} & 0.130 & \underline{0.990} & \underline{1.148} \\

        & \textbf{\ourmethod~(Ours)}
        & \textbf{5.385} & \textbf{34.894} & \textbf{0.520} & \textbf{0.056} & \textbf{0.273} & \textbf{1.571}
        & \textbf{3.341} & \textbf{12.932} & \textbf{0.592} & \textbf{0.075} & \textbf{0.312} & \textbf{0.981} \\

        \specialrule{.1em}{-0.05em}{-0.05em}
    \end{tabular}
    \vspace*{-1mm}
}
\end{center}
    \vspace*{-1.4em}
    \caption{
        \textbf{Quantitative comparison with state-of-the-art methods on Open4DHOI~\cite{wen2025efficient} and BEHAVE~\cite{bhatnagar2022behave}.}
        Bold and underlined values indicate the best and second-best scores, respectively.
    }
    \vspace*{-0.0em}
    \label{tab:sota_comparison}
\end{table*}

\subsection{Datasets}
We use Open4DHOI~\cite{wen2025efficient} and BEHAVE~\cite{bhatnagar2022behave} for our experiments.
Open4DHOI is an open-vocabulary, in-the-wild human-object interaction dataset spanning a wide range of object categories.
From the whole set, we curate a test set of 210 sequences, spanning diverse interactions.
BEHAVE is an indoor interaction dataset that captures the interactions of 8 human subjects and 20 objects in a controlled setting.
We use its official test split, consisting of 340 sequences.

\begin{figure}[t!]
  \centering
  \includegraphics[width=1.0\linewidth]{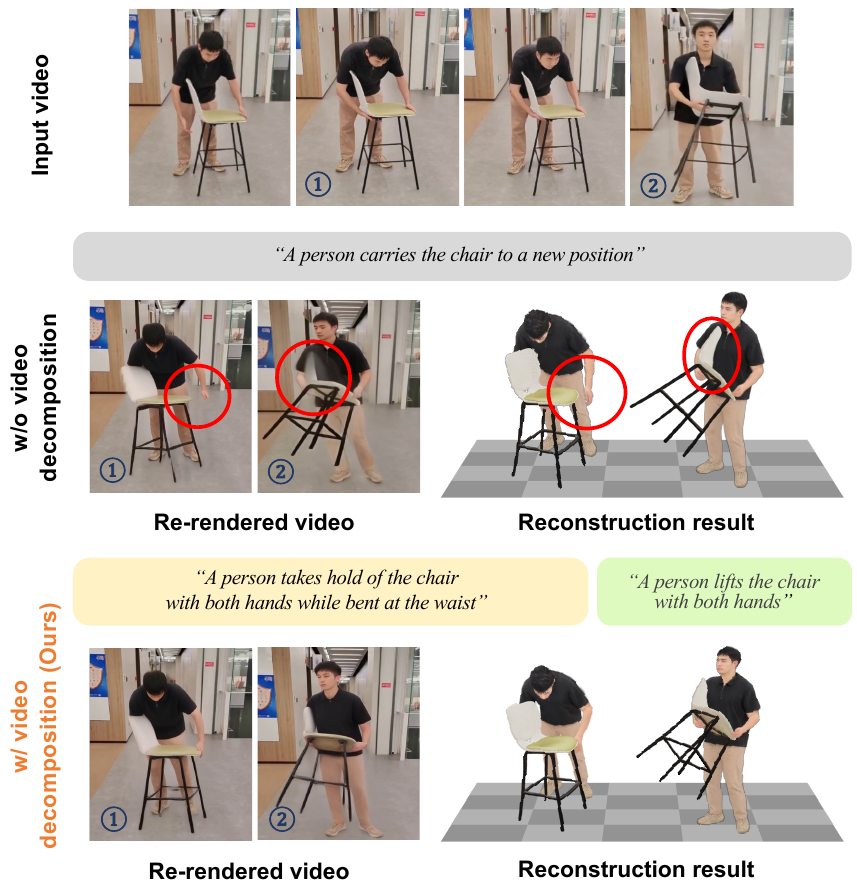}
  \vspace*{-1.5em}
  \caption{
  \textbf{Effectiveness of video decomposition.} 
  A single description of the entire video does not distinguish the interaction phases, resulting in implausible re-rendering and reconstruction. 
  Our decomposition assigns a description to each phase, guiding the re-rendering to follow the interaction at each moment.
  }
   \vspace*{+0.0em}
  \label{fig:abl_decomposition}
\end{figure}

\subsection{Evaluation metrics}
\label{sec:evaluation_metrics}
For all evaluations, we align the reconstruction to the ground truth (GT) per frame by translating the human root.

\noindent\textbf{Chamfer distance ($\text{CD}_{\text{H}}$, $\text{CD}_{\text{O}}$).}
We measure the Chamfer distance between the reconstructed human and object surfaces and their corresponding GT.
The Chamfer distance is separately computed for the human ($\text{CD}_{\text{H}}$) and object ($\text{CD}_{\text{O}}$) in centimeters.

\noindent\textbf{Contact score ($\text{Contact}$).}
We evaluate contact fidelity by comparing the contact regions of the reconstructed human and object against the GT.
We define the contact map as the human vertices of the SMPL-X mesh surface~\cite{pavlakos2019expressive} within $5~\text{cm}$ of the object surface.
The contact score ($\text{Contact}$) is computed as the harmonic mean of precision and recall between the predicted and GT human contact maps.

\noindent\textbf{Collision ratio ($\text{Collision}$).}
To evaluate physical plausibility, we measure the interpenetration between the reconstructed human and object surfaces.
The collision ratio ($\text{Collision}$) is the percentage of human vertices located within the object mesh.

\noindent\textbf{Acceleration error ($\text{Accel}_{\text{H}}$, $\text{Accel}_{\text{O}}$).}
To evaluate the temporal consistency of the reconstructed 4D motion, we measure the difference between the predicted and GT acceleration.
$\text{Accel}_{\text{H}}$ is computed over the 14 body joints of the SMPL-X model, and $\text{Accel}_{\text{O}}$ over the 3D object center.

\subsection{Ablation study}
We carry out all ablation studies on Open4DHOI~\cite{wen2025efficient}.

\noindent\textbf{Effectiveness of video decomposition.}
\cref{tab:abl_decomposition} and \cref{fig:abl_decomposition} show the effectiveness of our video decomposition with motion-aware captioning.
The first to third rows of \cref{tab:abl_decomposition} compare different captioning strategies, where standard captioning describes the interaction from the input video alone without object motion.
Using text captions (the second row) largely improves over no captioning (the first row) by providing semantic priors, and our motion-aware captioning (the third row) further improves the reconstruction by capturing dynamic movements from the object motion.
The third to fifth rows compare different decomposition strategies, where uniform decomposition splits the video into fixed 81-frame intervals.
A single global description (the third row) misses fine-grained interaction details, and uniform segments (the fourth row) often mix multiple interaction phases, yielding ambiguous text captions.
In contrast, our grounded decomposition (the fifth row) localizes the boundaries of each interaction phase, providing clear textual guidance that achieves the best results.
As shown in \cref{fig:abl_decomposition}, a single description of the entire video fails to distinguish holding the chair from lifting it, causing wrong reconstruction, whereas our decomposition guides the re-rendering to follow each interaction phase.

\noindent\textbf{Effectiveness of interaction-aware re-rendering.}
\cref{tab:abl_rerendering} shows the impact of our novel-view re-rendering pipeline and its text conditioning strategies.
Without novel-view re-rendering (the first row), reconstructing solely from monocular observations suffers from depth ambiguities, leading to wrong reconstruction.
In contrast, using re-rendering (the second row) provides multi-view supervision to mitigate depth ambiguities.
Furthermore, our regional text conditioning (the third row) guides the human and object with separate prompts, ensuring clear boundaries between them.
More qualitative re-rendering results, as in \cref{fig:1_introduction}, are provided in the supplementary material.

As described in \cref{sec:3.3_rerendering}, the re-rendering operates on each viewpoint independently, which may cause cross-view inconsistency between the re-rendered videos.
To prevent the inconsistency, our framework has three core designs.
First, the SDF loss $\mathcal{L}_{\text{SDF}}$ in 4D HOI optimization aggregates all viewpoints into a single target (\textit{i.e.}, SDF), which optimizes the object toward a geometric consensus across viewpoints.
Second, the re-rendering process is conditioned on the human skeleton (\cref{eq:refine}) projected from the same 4D human motion, which constrains the human-object interaction in every viewpoint to follow the underlying motion.
Third, the re-rendering runs on each short interaction phase obtained by the video decomposition, which alleviates the accumulation of cross-view discrepancies.
For a long video, each autoregressively generated view gradually deviates from the content generated in the other views, causing cross-view discrepancies.
Our framework reduces such discrepancies by re-rendering the short interaction phases obtained from the video decomposition.

\begin{figure}[t]
  \centering
  \vspace*{-1.5em}
  \includegraphics[width=1.0\linewidth]{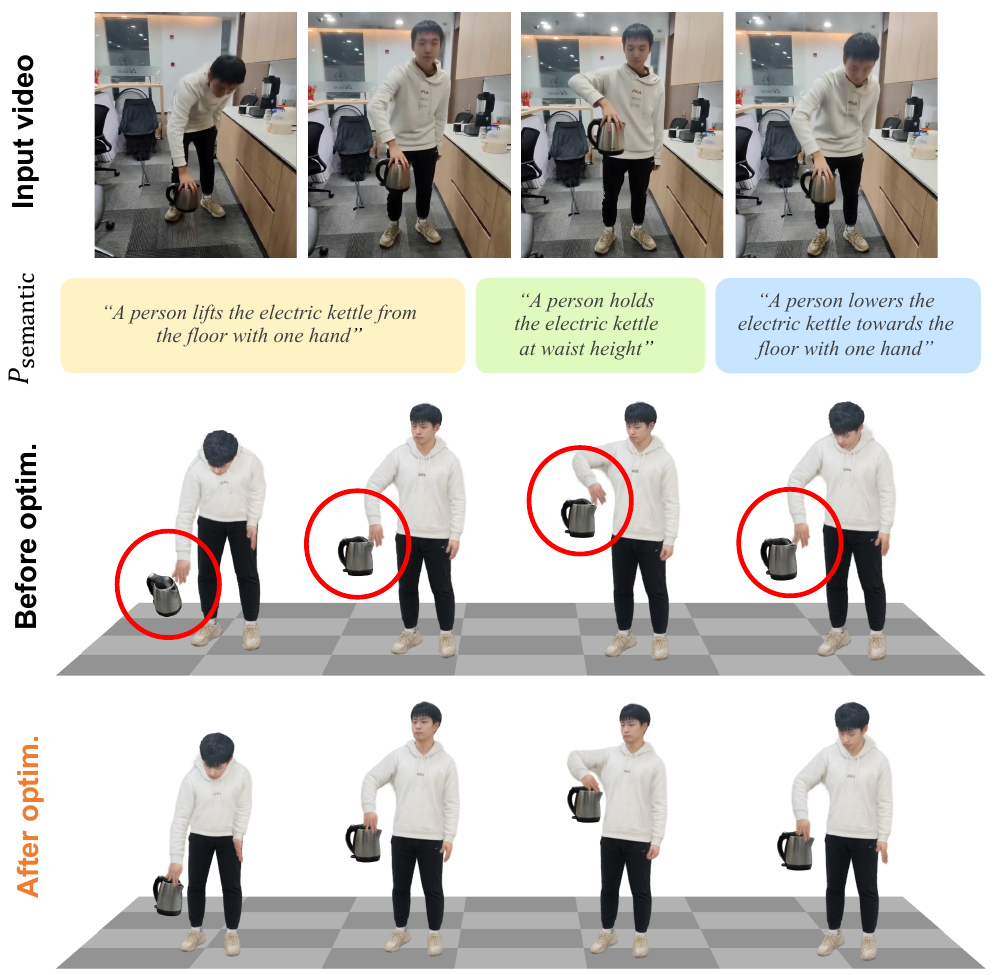}
  \vspace*{-1.3em}
  \caption{\textbf{Effectiveness of 4D HOI optimization.} 
  For each interaction phase described with $P_{\text{semantic}}$, our 4D HOI optimization corrects the artifacts (red circles) in the initial reconstructions, yielding accurate reconstruction results.
  }
   \vspace*{+0.0em}
  \label{fig:optimization_result}
\end{figure}

\begin{figure*}[t!]
  \centering
  \includegraphics[width=0.98\linewidth]{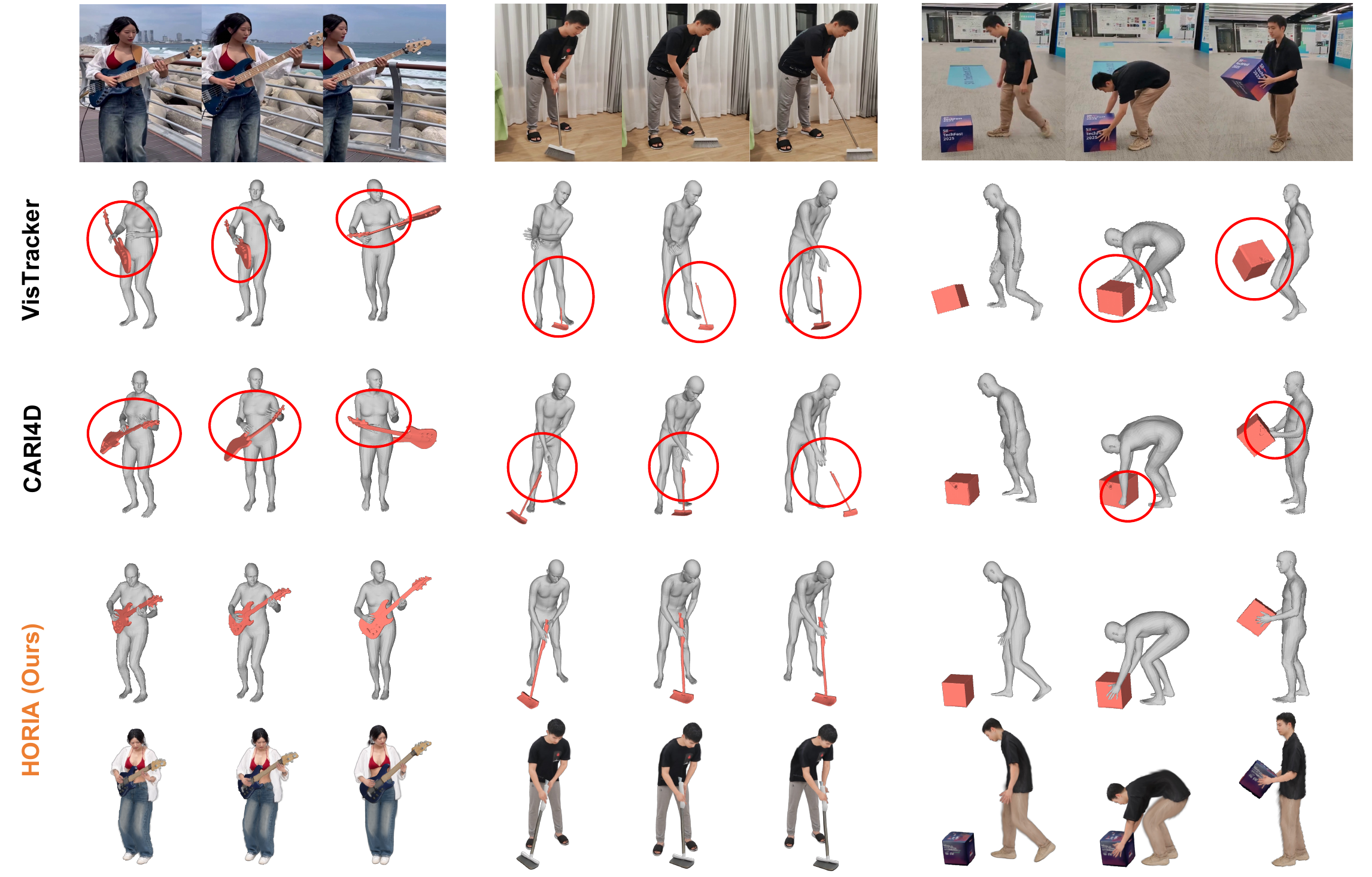}
  \vspace*{-0.1em}
  \caption{\textbf{Qualitative comparison with state-of-the-art methods.}
  We render each method's reconstruction at $-30^\circ$ azimuth and highlight representative failure cases with red circles.
  }
   \vspace*{+0.0em}
  \label{fig:6_sota_comparison}
\end{figure*}

\noindent\textbf{Ablation on 4D HOI optimization.}
\cref{fig:optimization_result} and \cref{tab:abl_optimization} show that our 4D HOI optimization effectively refines initial reconstructions using the re-rendered videos guided by the text descriptions.
As shown in \cref{fig:optimization_result}, the initial reconstruction in the first row of \cref{tab:abl_optimization} suffers from depth ambiguity and implausible artifacts.
Our 4D HOI optimization resolves these artifacts and yields temporally consistent motion by integrating auxiliary evidence from multi-view re-rendering in \cref{sec:3.3_rerendering}.
The second row shows that removing the SDF loss $\mathcal{L}_{\text{SDF}}$ largely degrades the object reconstruction.
Without $\mathcal{L}_{\text{SDF}}$, the re-rendered videos supervise the reconstruction only through $\mathcal{L}_{\text{recon}}$, which compares the rendering with each view independently and cannot resolve the inconsistencies across views.
$\mathcal{L}_{\text{SDF}}$ aggregates all viewpoints into a single 3D geometric consensus, guiding the object toward a geometry consistent across views.
The third row shows that the contact loss $\mathcal{L}_{\text{contact}}$ brings the human and object into physical contact, improving the contact score by a large margin.
The fourth row shows that the collision loss $\mathcal{L}_{\text{collision}}$ suppresses human-object interpenetration.

\subsection{Comparison with state-of-the-art methods}
We compare our framework with state-of-the-art reconstruction methods, categorized into image-based and video-based approaches.
Since the image-based methods optimize each frame independently, evaluating them on the full video is computationally expensive.
We thus evaluate the image-based methods on key frames sampled at 1\,fps, and report $\text{Accel}_{\text{H}}$ and $\text{Accel}_{\text{O}}$ only for the video-based methods.

\cref{tab:sota_comparison} and \cref{fig:6_sota_comparison} demonstrate that our framework outperforms all baselines both quantitatively and qualitatively on Open4DHOI~\cite{wen2025efficient} and BEHAVE~\cite{bhatnagar2022behave}.
Previous video-based methods frequently misplace the object relative to the hand or body during interaction, since they heavily rely on geometric constraints without semantic understanding of the interaction.
Unlike these methods, \ourmethod~leverages the text description of each interaction phase to capture the semantic context of the interaction.
Moreover, our interaction-aware re-rendering provides novel-view supervision, and the SDF loss enforces geometric consistency across these viewpoints, achieving the best performance.
\section{Conclusion}
\label{sec:conclusion}
We present \ourmethod, a framework for joint 4D reconstruction of a human and object from a monocular video.
Our framework decomposes the input video into several interaction phases and assigns a text description to each phase to capture fine-grained human-object interactions.
Guided by the text descriptions, our re-rendering refines the novel-view renderings of the reconstruction to compensate for limited single-view observations.
Extensive experiments show that \ourmethod~achieves state-of-the-art performance.

\clearpage
{
    \small
    \bibliographystyle{ieeenat_fullname}
    \bibliography{egbib}

@String(CVPR = {IEEE Conf. Comput. Vis. Pattern Recog.})

@String(ICCV = {Int. Conf. Comput. Vis.})

@String(ECCV = {Eur. Conf. Comput. Vis.})

@String(TOG  = {ACM Trans. Graph.})

@String(ICLR = {Int. Conf. Learn. Represent.})

@String(AAAI = {AAAI})

@inproceedings{pavlakos2019expressive,
  title={Expressive body capture: {3D} hands, face, and body from a single image},
  author={Pavlakos, Georgios and Choutas, Vasileios and Ghorbani, Nima and Bolkart, Timo and Osman, Ahmed AA and Tzionas, Dimitrios and Black, Michael J},
  booktitle={CVPR},
  year={2019}
}

@inproceedings{chen2019holistic++,
  title={Holistic++ scene understanding: Single-view {3D} holistic scene parsing and human pose estimation with human-object interaction and physical commonsense},
  author={Chen, Yixin and Huang, Siyuan and Yuan, Tao and Qi, Siyuan and Zhu, Yixin and Zhu, Song-Chun},
  booktitle={ICCV},
  year={2019}
}

@inproceedings{zhang2020perceiving,
  title={Perceiving {3D} human-object spatial arrangements from a single image in the wild},
  author={Zhang, Jason Y and Pepose, Sam and Joo, Hanbyul and Ramanan, Deva and Malik, Jitendra and Kanazawa, Angjoo},
  booktitle={ECCV},
  year={2020}
}

@article{xu2021d3d,
  title={{D3D-HOI}: Dynamic {3D} human-object interactions from videos},
  author={Xu, Xiang and Joo, Hanbyul and Mori, Greg and Savva, Manolis},
  journal={arXiv preprint arXiv:2108.08420},
  year={2021}
}

@inproceedings{xie2022chore,
  title={{CHORE}: Contact, human and object reconstruction from a single {RGB} image},
  author={Xie, Xianghui and Bhatnagar, Bharat Lal and Pons-Moll, Gerard},
  booktitle={ECCV},
  year={2022},
}

@inproceedings{bhatnagar2022behave,
  title={{BEHAVE}: Dataset and method for tracking human object interactions},
  author={Bhatnagar, Bharat Lal and Xie, Xianghui and Petrov, Ilya A and Sminchisescu, Cristian and Theobalt, Christian and Pons-Moll, Gerard},
  booktitle={CVPR},
  year={2022}
}

@inproceedings{wen2025reconstructing,
  title={Reconstructing in-the-wild open-vocabulary human-object interactions},
  author={Wen, Boran and Huang, Dingbang and Zhang, Zichen and Zhou, Jiahong and Deng, Jianbin and Gong, Jingyu and Chen, Yulong and Ma, Lizhuang and Li, Yong-Lu},
  booktitle={CVPR},
  year={2025}
}

@inproceedings{nam2026tehor,
  title={{TeHOR}: Text-Guided {3D} Human and Object Reconstruction with Textures},
  author={Nam, Hyeongjin and Jung, Daniel Sungho and Lee, Kyoung Mu},
  booktitle={CVPR},
  year={2026}
}

@article{wen2025efficient,
  title={Efficient and Scalable Monocular Human-Object Interaction Motion Reconstruction},
  author={Wen, Boran and Lu, Ye and Wang, Sirui and Wan, Keyan and Zhou, Jiahong and Liang, Junxuan and Liu, Xinpeng and Xiao, Bang and Liu, Ruiyang and Li, Yong-Lu},
  journal={arXiv preprint arXiv:2512.00960},
  year={2025}
}

@inproceedings{xie2026cari4d,
    title = {{CARI4D}: Category Agnostic {4D} Reconstruction of Human-Object Interaction},
    author = {Xie, Xianghui and Wen, Bowen and Chang, Yan and Rabeti, Hesam and Li, Jiefeng and Yuan, Ye and Pons-Moll, Gerard and Birchfield, Stan},
    booktitle = {CVPR},
    year = {2026},
}

@inproceedings{xie2023templatefree,
  title={Template Free Reconstruction of Human-object Interaction with Procedural Interaction Generation},
  author={Xie, Xianghui and Bhatnagar, Bharat Lal and Lenssen, Jan Eric and Pons-Moll, Gerard},
  booktitle={CVPR},
  year={2024},
}

@inproceedings{dwivedi2025interactvlm,
  title={{InteractVLM}: {3D} interaction reasoning from {2D} foundational models},
  author={Dwivedi, Sai Kumar and Anti{\'c}, Dimitrije and Tripathi, Shashank and Taheri, Omid and Schmid, Cordelia and Black, Michael J and Tzionas, Dimitrios},
  booktitle={CVPR},
  year={2025}
}

@inproceedings{nam2024contho,    
    title={Joint Reconstruction of {3D} Human and Object via Contact-Based Refinement Transformer},
    author={Nam, Hyeongjin and Jung, Daniel Sungho and Moon, Gyeongsik and Lee, Kyoung Mu},
    booktitle={CVPR},  
    year={2024}  
}

@inproceedings{li2025scorehoi,
  title={{ScoreHOI}: Physically Plausible Reconstruction of Human-Object Interaction via Score-Guided Diffusion},
  author={Li, Ao and Liu, Jinpeng and Zhu, Yixuan and Tang, Yansong},
  booktitle={ICCV},
  year={2025}
}

@inproceedings{xie2025intertrack,
  title={{InterTrack}: Tracking human object interaction without object templates},
  author={Xie, Xianghui and Lenssen, Jan Eric and Pons-Moll, Gerard},
  booktitle={3DV},
  year={2025},
}

@inproceedings{xie2023visibility,
  title={Visibility aware human-object interaction tracking from single {RGB} camera},
  author={Xie, Xianghui and Bhatnagar, Bharat Lal and Pons-Moll, Gerard},
  booktitle={CVPR},
  year={2023}
}

@inproceedings{huang2026recovering,
  title={Recovering Physically Plausible Human-Object Interactions from Monocular Videos},
  author={Huang, Dingbang and Vouga, Etienne and Huang, Qixing and Pavlakos, Georgios},
  booktitle={CVPR},
  year={2026}
}

@inproceedings{xue2026rhino,
  title={{RHINO}: Reconstructing Human Interactions with Novel Objects from Monocular Videos},
  author={Xue, Lixin and Zheng, Chengwei and Paschalidis, Georgios and Guo, Chen and Kaufmann, Manuel and Zarate, Juan and Tzionas, Dimitrios},
  booktitle={CVPR},
  year={2026}
}

@book{de1978practical,
  title={A practical guide to splines},
  author={De Boor, Carl},
  year={1978},
  publisher={Springer New York}
}

@inproceedings{huang2025vivid4d,
  title={{Vivid4D}: Improving {4D} reconstruction from monocular video by video inpainting},
  author={Huang, Jiaxin and Miao, Sheng and Yang, Bangbang and Ma, Yuewen and Liao, Yiyi},
  booktitle={ICCV},
  year={2025}
}

@article{openai2026gpt5,
  title={{GPT-5} system card},
  author={OpenAI},
  journal={arXiv preprint arXiv:2601.03267},
  year={2026}
}

@article{bai2025qwen3,
  title={{Qwen3-VL} technical report},
  author={Bai, Shuai and Cai, Yuxuan and Chen, Ruizhe and Chen, Keqin and Chen, Xionghui and Cheng, Zesen and Deng, Lianghao and Ding, Wei and Gao, Chang and Ge, Chunjiang and others},
  journal={arXiv preprint arXiv:2511.21631},
  year={2025}
}

@article{wan2025wan,
  title={{Wan}: Open and advanced large-scale video generative models},
  author={Wan, Team and Wang, Ang and Ai, Baole and Wen, Bin and Mao, Chaojie and Xie, Chen-Wei and Chen, Di and Yu, Feiwu and Zhao, Haiming and Yang, Jianxiao and others},
  journal={arXiv preprint arXiv:2503.20314},
  year={2025}
}

@inproceedings{huang2024tech,
  title={{{TeCH}: Text-guided Reconstruction of Lifelike Clothed Humans}},
  author={Huang, Yangyi and Yi, Hongwei and Xiu, Yuliang and Liao, Tingting and Tang, Jiaxiang and Cai, Deng and Thies, Justus},
  booktitle={3DV},
  year={2024}
}

@inproceedings{tang2024dreamgaussian,
  title={{DreamGaussian}: Generative gaussian splatting for efficient {3D} content creation},
  author={Tang, Jiaxiang and Ren, Jiawei and Zhou, Hang and Liu, Ziwei and Zeng, Gang},
  booktitle={ICLR},
  year={2024}
}

@inproceedings{qian2024magic123,
  title={{Magic123}: One image to high-quality {3D} object generation using both {2D} and {3D} diffusion priors},
  author={Qian, Guocheng and Mai, Jinjie and Hamdi, Abdullah and Ren, Jian and Siarohin, Aliaksandr and Li, Bing and Lee, Hsin-Ying and Skorokhodov, Ivan and Wonka, Peter and Tulyakov, Sergey and others},
  booktitle={ICLR},
  year={2024}
}

@inproceedings{tang2023make,
  title={{Make-it-3D}: High-fidelity {3D} creation from a single image with diffusion prior},
  author={Tang, Junshu and Wang, Tengfei and Zhang, Bo and Zhang, Ting and Yi, Ran and Ma, Lizhuang and Chen, Dong},
  booktitle={ICCV},
  year={2023},
}

@inproceedings{ye2024ghop,
  title={{G-HOP}: Generative Hand-Object Prior for Interaction Reconstruction and Grasp Synthesis},
  author={Ye, Yufei and Gupta, Abhinav and Kitani, Kris and Tulsiani, Shubham},
  booktitle={CVPR},
  year={2024}
}

@inproceedings{wang2024animatabledreamer,
  title={{AnimatableDreamer}: Text-Guided Non-rigid {3D} Model Generation and Reconstruction with Canonical Score Distillation},
  author={Wang, Xinzhou and Wang, Yikai and Ye, Junliang and Sun, Fuchun and Wang, Zhengyi and Wang, Ling and Liu, Pengkun and Sun, Kai and Wang, Xintong and Xie, Wende and Liu, Fangfu and He, Bin},
  booktitle={ECCV},
  year={2024}
}

@inproceedings{deng2023nerdi,
  title={{NeRDi}: Single-view {NeRF} synthesis with language-guided diffusion as general image priors},
  author={Deng, Congyue and Jiang, Chiyu and Qi, Charles R and Yan, Xinchen and Zhou, Yin and Guibas, Leonidas and Anguelov, Dragomir and others},
  booktitle={CVPR},
  year={2023}
}

@inproceedings{sun2024dreamcraft3d,
  title={{DreamCraft3D}: Hierarchical {3D} generation with bootstrapped diffusion prior},
  author={Sun, Jingxiang and Zhang, Bo and Shao, Ruizhi and Wang, Lizhen and Liu, Wen and Xie, Zhenda and Liu, Yebin},
  booktitle={ICLR},
  year={2024}
}

@inproceedings{rombach2022high,
  title={High-resolution image synthesis with latent diffusion models},
  author={Rombach, Robin and Blattmann, Andreas and Lorenz, Dominik and Esser, Patrick and Ommer, Bj{\"o}rn},
  booktitle={CVPR},
  year={2022}
}

@article{kerbl3Dgaussians,
  author={Kerbl, Bernhard and Kopanas, Georgios and Leimk{\"u}hler, Thomas and Drettakis, George},
  title={{3D} gaussian splatting for real-time radiance field rendering},
  journal={ACM TOG},
  year={2023}
}

@inproceedings{yu2024mip,
  title={{Mip-Splatting}: Alias-free {3D} gaussian splatting},
  author={Yu, Zehao and Chen, Anpei and Huang, Binbin and Sattler, Torsten and Geiger, Andreas},
  booktitle={CVPR},
  year={2024}
}

@inproceedings{qiu2025lhm,
  title={{LHM}: Large animatable human reconstruction model from a single image in seconds},
  author={Qiu, Lingteng and Gu, Xiaodong and Li, Peihao and Zuo, Qi and Shen, Weichao and Zhang, Junfei and Qiu, Kejie and Yuan, Weihao and Chen, Guanying and Dong, Zilong and others},
  booktitle={ICCV},
  year={2025}
}

@inproceedings{moon2024expressive,
  title={Expressive whole-body {3D} gaussian avatar},
  author={Moon, Gyeongsik and Shiratori, Takaaki and Saito, Shunsuke},
  booktitle={ECCV},
  year={2024}
}

@inproceedings{chen2026sam,
  title={{SAM 3D}: {3Dfy} anything in images},
  author={Chen, Xingyu and Chu, Fu-Jen and Gleize, Pierre and Liang, Kevin J and Sax, Alexander and Tang, Hao and Wang, Weiyao and Guo, Michelle and Hardin, Thibaut and Li, Xiang and others},
  booktitle={CVPR},
  year={2026}
}

@inproceedings{zhou2019continuity,
  title={On the continuity of rotation representations in neural networks},
  author={Zhou, Yi and Barnes, Connelly and Lu, Jingwan and Yang, Jimei and Li, Hao},
  booktitle={CVPR},
  year={2019},
}

@inproceedings{baradel2024multi,
  title={{Multi-HMR}: Multi-person whole-body human mesh recovery in a single shot},
  author={Baradel, Fabien and Armando, Matthieu and Galaaoui, Salma and Br{\'e}gier, Romain and Weinzaepfel, Philippe and Rogez, Gr{\'e}gory and Lucas, Thomas},
  booktitle={ECCV},
  year={2024}
}

@inproceedings{wen2024foundationpose,
  title={{FoundationPose}: Unified {6D} pose estimation and tracking of novel objects},
  author={Wen, Bowen and Yang, Wei and Kautz, Jan and Birchfield, Stan},
  booktitle={CVPR},
  year={2024},
}

@article{douglas1973algorithms,
  title={Algorithms for the reduction of the number of points required to represent a digitized line or its caricature},
  author={Douglas, David H and Peucker, Thomas K},
  journal={Cartographica},
  year={1973},
}

@article{zhang2026timelens,
  title={{TimeLens}: Rethinking Video Temporal Grounding with Multimodal {LLMs}},
  author={Zhang, Jun and Wang, Teng and Ge, Yuying and Ge, Yixiao and Li, Xinhao and Shan, Ying and Wang, Limin},
  journal={CVPR},
  year={2026}
}

@inproceedings{jiang2025smarteraser,
  title={{SmartEraser}: Remove anything from images using masked-region guidance},
  author={Jiang, Longtao and Wang, Zhendong and Bao, Jianmin and Zhou, Wengang and Chen, Dongdong and Shi, Lei and Chen, Dong and Li, Houqiang},
  booktitle={CVPR},
  year={2025}
}

@inproceedings{lipman2023flow,
  title={Flow matching for generative modeling},
  author={Lipman, Yaron and Chen, Ricky TQ and Ben-Hamu, Heli and Nickel, Maximilian and Le, Matt},
  booktitle={ICLR},
  year={2023}
}

@inproceedings{ravi2025sam,
  title={{SAM 2}: Segment anything in images and videos},
  author={Ravi, Nikhila and Gabeur, Valentin and Hu, Yuan-Ting and Hu, Ronghang and Ryali, Chaitanya and Ma, Tengyu and Khedr, Haitham and R{\"a}dle, Roman and Rolland, Chloe and Gustafson, Laura and others},
  booktitle={ICLR},
  year={2025}
}

@inproceedings{piccinelli2024unidepth,
  title={{UniDepth}: Universal monocular metric depth estimation},
  author={Piccinelli, Luigi and Yang, Yung-Hsu and Sakaridis, Christos and Segu, Mattia and Li, Siyuan and Van Gool, Luc and Yu, Fisher},
  booktitle={CVPR},
  year={2024},
}

@inproceedings{ester1996density,
  title={A Density-Based Algorithm for Discovering Clusters in Large Spatial Databases with Noise},
  author={Ester, Martin and Kriegel, Hans-Peter and Sander, J{\"o}rg and Xu, Xiaowei},
  booktitle={KDD},
  year={1996}
}

@inproceedings{jiang2020coherent,
  title={Coherent reconstruction of multiple humans from a single image},
  author={Jiang, Wen and Kolotouros, Nikos and Pavlakos, Georgios and Zhou, Xiaowei and Daniilidis, Kostas},
  booktitle={CVPR},
  year={2020}
}

@inproceedings{gao2017tall,
  title={{TALL}: Temporal activity localization via language query},
  author={Gao, Jiyang and Sun, Chen and Yang, Zhenheng and Nevatia, Ram},
  booktitle={ICCV},
  year={2017},
}

@inproceedings{zhang2020learning,
  title={Learning {2D} temporal adjacent networks for moment localization with natural language},
  author={Zhang, Songyang and Peng, Houwen and Fu, Jianlong and Luo, Jiebo},
  booktitle={AAAI},
  year={2020}
}

@inproceedings{lei2021moment,
  title={{QVHighlights}: Detecting Moments and Highlights in Videos via Natural Language Queries},
  author={Lei, Jie and Berg, Tamara L and Bansal, Mohit},
  booktitle={NeurIPS},
  year={2021}
}

@inproceedings{moon2023query,
  title={Query-dependent video representation for moment retrieval and highlight detection},
  author={Moon, WonJun and Hyun, Sangeek and Park, SangUk and Park, Dongchan and Heo, Jae-Pil},
  booktitle={CVPR},
  year={2023}
}

@inproceedings{lin2023univtg,
  title={{UniVTG}: Towards unified video-language temporal grounding},
  author={Lin, Kevin Qinghong and Zhang, Pengchuan and Chen, Joya and Pramanick, Shraman and Gao, Difei and Wang, Alex Jinpeng and Yan, Rui and Shou, Mike Zheng},
  booktitle={ICCV},
  year={2023},
}

@inproceedings{ren2024timechat,
  title={{TimeChat}: A time-sensitive multimodal large language model for long video understanding},
  author={Ren, Shuhuai and Yao, Linli and Li, Shicheng and Sun, Xu and Hou, Lu},
  booktitle={CVPR},
  year={2024}
}

@inproceedings{huang2024vtimellm,
  title={{VTimeLLM}: Empower {LLM} to grasp video moments},
  author={Huang, Bin and Wang, Xin and Chen, Hong and Song, Zihan and Zhu, Wenwu},
  booktitle={CVPR},
  year={2024},
}

@article{wang2024groundedvideollm,
  title={{Grounded-VideoLLM}: Sharpening Fine-grained Temporal Grounding in Video Large Language Models},
  author={Wang, Haibo and Xu, Zhiyang and Cheng, Yu and Diao, Shizhe and Zhou, Yufan and Cao, Yixin and Wang, Qifan and Ge, Weifeng and Huang, Lifu},
  journal={arXiv preprint arXiv:2410.03290},
  year={2024}
}

@article{chen2025fmimic,
  title={{FMimic}: Foundation models are fine-grained action learners from human videos},
  author={Chen, Guangyan and Wang, Meiling and Cui, Te and Mu, Yao and Lu, Haoyang and Peng, Zicai and Hu, Mengxiao and Zhou, Tianxing and Fu, Mengyin and Yang, Yi and others},
  journal={IJRR},
  year={2025},
}

@inproceedings{vaswani2017attention,
  title={Attention is all you need},
  author={Vaswani, Ashish and Shazeer, Noam and Parmar, Niki and Uszkoreit, Jakob and Jones, Llion and Gomez, Aidan N and Kaiser, {\L}ukasz and Polosukhin, Illia},
  booktitle={NeurIPS},
  year={2017}
}

@inproceedings{poole2023dreamfusion,
  author = {Poole, Ben and Jain, Ajay and Barron, Jonathan T. and Mildenhall, Ben},
  title = {{DreamFusion}: Text-to-{3D} using {2D} Diffusion},
  booktitle = {ICLR},
  year = {2023},
}

@inproceedings{wang2023prolificdreamer,
  title={{ProlificDreamer}: High-fidelity and diverse text-to-{3D} generation with variational score distillation},
  author={Wang, Zhengyi and Lu, Cheng and Wang, Yikai and Bao, Fan and Li, Chongxuan and Su, Hang and Zhu, Jun},
  booktitle={NeurIPS},
  year={2023}
}

@inproceedings{huang2022intercap,
  title={{InterCap}: Joint Markerless {3D} Tracking of Humans and Objects in Interaction},
  author={Huang, Yinghao and Taheri, Omid and Black, Michael J and Tzionas, Dimitrios},
  booktitle={GCPR},
  year={2022},
}

@inproceedings{song2021score,
  title={Score-based generative modeling through stochastic differential equations},
  author={Song, Yang and Sohl-Dickstein, Jascha and Kingma, Diederik P and Kumar, Abhishek and Ermon, Stefano and Poole, Ben},
  booktitle={ICLR},
  year={2021}
}

@inproceedings{kingma2015adam,
    title={{Adam}: A Method for Stochastic Optimization}, 
    author={Diederik P. Kingma and Jimmy Ba},
    booktitle = {ICLR},
    year={2015},
}

@inproceedings{cseke2025pico,
    title= {{PICO}: Reconstructing {3D} People In Contact with Objects},
    author={Cseke, Alp\'{a}r and Tripathi, Shashank and Dwivedi, Sai Kumar and Lakshmipathy, Arjun and Chatterjee, Agniv and Black, Michael J. and Tzionas, Dimitrios},
    booktitle={CVPR},
    year={2025},
}

@inproceedings{liu2023zero,
  title={{Zero-1-to-3}: Zero-shot one image to {3D} object},
  author={Liu, Ruoshi and Wu, Rundi and Van Hoorick, Basile and Tokmakov, Pavel and Zakharov, Sergey and Vondrick, Carl},
  booktitle={ICCV},
  year={2023},
}

@inproceedings{radford2021learning,
  title={Learning transferable visual models from natural language supervision},
  author={Radford, Alec and Kim, Jong Wook and Hallacy, Chris and Ramesh, Aditya and Goh, Gabriel and Agarwal, Sandhini and Sastry, Girish and Askell, Amanda and Mishkin, Pamela and Clark, Jack and others},
  booktitle={ICML},
  year={2021},
}
}

\clearpage
\setcounter{page}{1}
\twocolumn[{%
\centering
\Large
\textbf{\thetitle}\\
\vspace{0.5em}Supplementary Material \\
\vspace{2.5em}
}]

\setcounter{section}{0}
\setcounter{table}{0}
\setcounter{figure}{0}
\renewcommand{\thesection}{S\arabic{section}}   
\renewcommand{\thetable}{S\arabic{table}}   
\renewcommand{\thefigure}{S\arabic{figure}}

In this supplementary material, we present additional technical details and more experimental results that could not be included in the main manuscript due to the lack of pages.
The contents are summarized below:
\begin{itemize}
\vspace{2.5mm}
\item \ref{sec:suppl_intercap}. Comparison on InterCap dataset
\item \ref{sec:suppl_implementation}. Implementation details
\item \ref{sec:suppl_decomp_eval}. Evaluation of video decomposition
\item \ref{sec:suppl_ablation}. More ablation studies
\item \ref{sec:suppl_runtime}. Comparison of running time
\item \ref{sec:suppl_limitation}. Limitations and future work
\item \ref{sec:suppl_comparison}. More comparisons with state-of-the-art methods
\end{itemize}


\section{Comparison on InterCap}
\label{sec:suppl_intercap}
\cref{tab:suppl_intercap} shows that \ourmethod~also outperforms existing reconstruction methods on the InterCap~\cite{huang2022intercap} dataset.
Similar to BEHAVE~\cite{bhatnagar2022behave}, InterCap is an indoor human-object interaction dataset that captures 10 subjects interacting with 10 rigid objects, including whole-body motion with the hands.
We use its official test split, consisting of 240 sequences, for evaluation.
Compared to both image-based and video-based methods, \ourmethod~achieves more accurate reconstruction with fewer interpenetrations.

\section{Implementation details}
\label{sec:suppl_implementation}
We provide the implementation details of three stages: video decomposition (Sec.~\textcolor{cvprblue}{3.2}), interaction-aware re-rendering (Sec.~\textcolor{cvprblue}{3.3}), and 4D HOI optimization (Sec.~\textcolor{cvprblue}{3.4}).

\subsection{Video decomposition}
\label{sec:suppl_decomposition}

\noindent\textbf{Coarse boundary detection.}
To detect initial segment boundaries where the object trajectory changes sharply, we apply the RDP algorithm~\cite{douglas1973algorithms} to the relative trajectory between the object and human root.
Before running RDP, we smooth this relative trajectory using a moving average filter with a window size of 5 frames to remove the estimation noise of the initial object motion.
The RDP algorithm then splits the video where this relative trajectory deviates by more than 10\,cm from the simplified linear trajectory.
To prevent overly short segments, we set the minimum length of each split segment to 15 frames.

\begin{table}[t]
\def\arraystretch{1.6}
\renewcommand{\tabcolsep}{0.6mm}
\dashlinedash 1.5pt
\dashlinegap 2.0pt
\footnotesize
\begin{center}
\scalebox{0.7}{
    \begin{tabular}{
    >{\centering\arraybackslash}m{0.4cm}
    >{\raggedright\arraybackslash}m{2.65cm}|
    >{\centering\arraybackslash}m{1.2cm}
    >{\centering\arraybackslash}m{1.2cm}
    >{\centering\arraybackslash}m{1.35cm}
    >{\centering\arraybackslash}m{1.5cm}
    >{\centering\arraybackslash}m{1.3cm}
    >{\centering\arraybackslash}m{1.3cm}
    }
    \specialrule{.1em}{.05em}{0.0em}

        \multicolumn{2}{l|}{Methods}
        & $\text{CD}_{\text{H}}{\downarrow}$
        & $\text{CD}_{\text{O}}{\downarrow}$
        & $\text{Contact}{\uparrow}$
        & $\text{Collision}{\downarrow}$
        & $\text{Accel}_{\text{H}}{\downarrow}$
        & $\text{Accel}_{\text{O}}{\downarrow}$ \\
        \hline

        \multirow{5}{*}{\rotatebox{90}{Image-based\;\;}}
        & PHOSA~\cite{zhang2020perceiving}
        & 4.622 & 42.161 & 0.375 & 0.079 & -- & -- \\

        & HOI-Gaussian~\cite{wen2025reconstructing}
        & 2.891 & 39.542 & 0.509 & 0.080 & -- & -- \\

        & TeHOR~\cite{nam2026tehor}
        & \underline{2.528} & \underline{32.160} & \underline{0.513} & \underline{0.073} & -- & -- \\
        \cdashline{2-8}

        & \rule{0pt}{2.3ex}\ourmethod~(Init.)
        & 2.705 & 63.394 & 0.406 & 0.098 & -- & -- \\

        & \textbf{\ourmethod~(Ours)}
        & \textbf{2.524} & \textbf{14.951} & \textbf{0.543} & \textbf{0.069} & -- & -- \\

        \hline

        \multirow{5}{*}{\rotatebox{90}{Video-based\;\;}}
        & VisTracker~\cite{xie2023visibility}
        & 6.152 & 23.257 & 0.405 & 0.105 & 1.216 & 2.781 \\

        & InterTrack~\cite{xie2025intertrack}
        & 10.019 & 20.959 & 0.487 & 0.218 & 1.985 & 4.104 \\

        & CARI4D~\cite{xie2026cari4d}
        & \textbf{2.415} & \underline{16.741} & \underline{0.516} & \underline{0.084} & 1.047 & \underline{2.129} \\
        \cdashline{2-8}

        & \rule{0pt}{2.3ex}\ourmethod~(Init.)
        & 2.728 & 62.777 & 0.402 & 0.101 & \underline{0.825} & 21.712 \\

        & \textbf{\ourmethod~(Ours)}
        & \underline{2.501} & \textbf{14.551} & \textbf{0.539} & \textbf{0.069} & \textbf{0.300} & \textbf{0.942} \\

        \specialrule{.1em}{-0.05em}{-0.05em}
    \end{tabular}
    \vspace*{-1mm}
}
\end{center}
    \vspace*{-1.4em}
    \caption{
        \textbf{Quantitative comparison with state-of-the-art methods on InterCap~\cite{huang2022intercap}.}
        Bold and underlined values indicate the best and second-best scores, respectively.
    }
    \vspace*{-0.0em}
    \label{tab:suppl_intercap}
\end{table}

\noindent\textbf{Motion-aware captioning.}
We extract the semantic prompt $P_{\text{semantic}}$ and contact prompt $P_{\text{contact}}$ for each coarse segment using a VLM (\textit{i.e.}, GPT-5~\cite{openai2026gpt5}) with the text instructions shown in \cref{fig:suppl_prompt_semantic,fig:suppl_prompt_contact}.
For each segment, we uniformly sample five frames, resize their longer side to 224 pixels, and feed them into the VLM.
To inform the VLM of how the object moves relative to the human, we provide a motion hint, shown in the lower part of \cref{fig:suppl_prompt_semantic}.
It describes the 3D distance change between the object and seven body parts: head, chest, hips, left and right hands, and left and right feet.
Comparing the first and last frames of the segment with a tolerance of $\delta=10$\,cm, we categorize the distance change of each part as \textit{moves closer} when the distance decreases by at least $\delta$, \textit{moves away} when it increases by at least $\delta$, and \textit{unchanged} otherwise.
For $P_{\text{contact}}$ (\cref{fig:suppl_prompt_contact}), the VLM selects the body parts in physical contact with the object from 18 pre-defined candidates, which are then mapped to the corresponding SMPL-X~\cite{pavlakos2019expressive} body vertex groups.

\begin{figure}[t]
    \begin{lstlisting}[]
You are a precise video annotator for hand-object interaction segments.
Provide a detailed description of the human-object interaction shown across these uniformly sampled frames of the given video.

The interacted object is (*@\textbf{\{object\}}@*).
Here is information about changes in the distance between the object and each body part. 
These measurements may contain errors; use them as supplementary reference information.
Hint: (*@\textbf{\{motion hint\}}@*)

Task:
- Write exactly one short English caption describing the dominant human-object interaction across these frames.
- Refer to the object as "(*@\textbf{\{object\}}@*)".
- Prefer a motion or state change over a static appearance description.
- Express exactly one action; do not use the words "and" or "then".
- Do not include timestamp phrases, frame numbers, or vague statements.

Output Format: {"caption": str}
    \end{lstlisting}
    \begin{lstlisting}[]
(*@$\triangleright$ \textbf{\{Motion hint\}}@*) (example)

3D distance changes between the object and each body part from the first to last frame of this segment:
- Head: moves closer
- Chest: moves closer
- Hips: unchanged
- Left hand: unchanged
- Right hand: moves closer
- Left foot: moves away
- Right foot: moves away
    \end{lstlisting}
    \vspace{-0.8em}
\caption{\textbf{Motion-aware captioning instruction for $P_{\text{semantic}}$.} 
The lower box shows an example of the object motion hint, which summarizes how the distance between the object and each body part changes within the segment.}
\label{fig:suppl_prompt_semantic}
\end{figure}

\noindent\textbf{Fine boundary grounding.}
To refine the coarse segments into precise interaction phases, we ground each caption across the full video using TimeLens-7B~\cite{zhang2026timelens}.
For each $P_{\text{semantic}}$ obtained from the motion-aware captioning, we query TimeLens with the full video sampled at 6\,fps, and it predicts the start and end times of the described interaction.
After the grounding, we merge consecutive predicted intervals whose temporal intersection over union reaches $0.9$.
The merged intervals serve as the final interaction phases for the re-rendering.

\subsection{Interaction-aware re-rendering}
\label{sec:suppl_impl_rerendering}

\noindent\textbf{Novel-view rendering.}
To obtain the novel-view videos for each interaction phase, we place six novel-view cameras at $\{-90^{\circ}, -60^{\circ}, -30^{\circ}, +30^{\circ}, +60^{\circ}, +90^{\circ}\}$ from the viewpoint of the input video.
These novel-view cameras are centered at the average 3D position of the human root and the object in the first frame.
For each viewpoint, we render the 3D human and object Gaussians posed by their initial motions onto the 2D background at a fixed resolution of $480 \times 832$.

\begin{figure}[t]
    \begin{lstlisting}[]
You are a precise video annotator for hand-object interaction segments.
Provide the contact information for the human-object interaction shown across these uniformly sampled frames of the given video.

The interacted object is (*@\textbf{\{object\}}@*).

Task:
- Identify the body parts that are in direct physical contact with the object during this chunk.
- Choose only from this pre-defined list; multiple parts are allowed: head, torso, hips, ...
- Focus only on the person whose body center is closest to the image center.
- "left" and "right" are the person's anatomical sides, not the viewer's sides.
- For no visible contact, return: {"contact_parts": []}

Examples:
- {"contact_parts": ["right hand"]}
- {"contact_parts": ["right arm", "right elbow"]}
- {"contact_parts": []}

Output Format: {"contact_parts": ["part name", ...]}
    \end{lstlisting}
    \vspace{-0.8em}
\caption{\textbf{Motion-aware captioning instruction for $P_{\text{contact}}$.}}
\label{fig:suppl_prompt_contact}
    \vspace{-0.4em}
\end{figure}

\begin{figure*}[t!]
  \centering
  \includegraphics[width=0.98\linewidth]{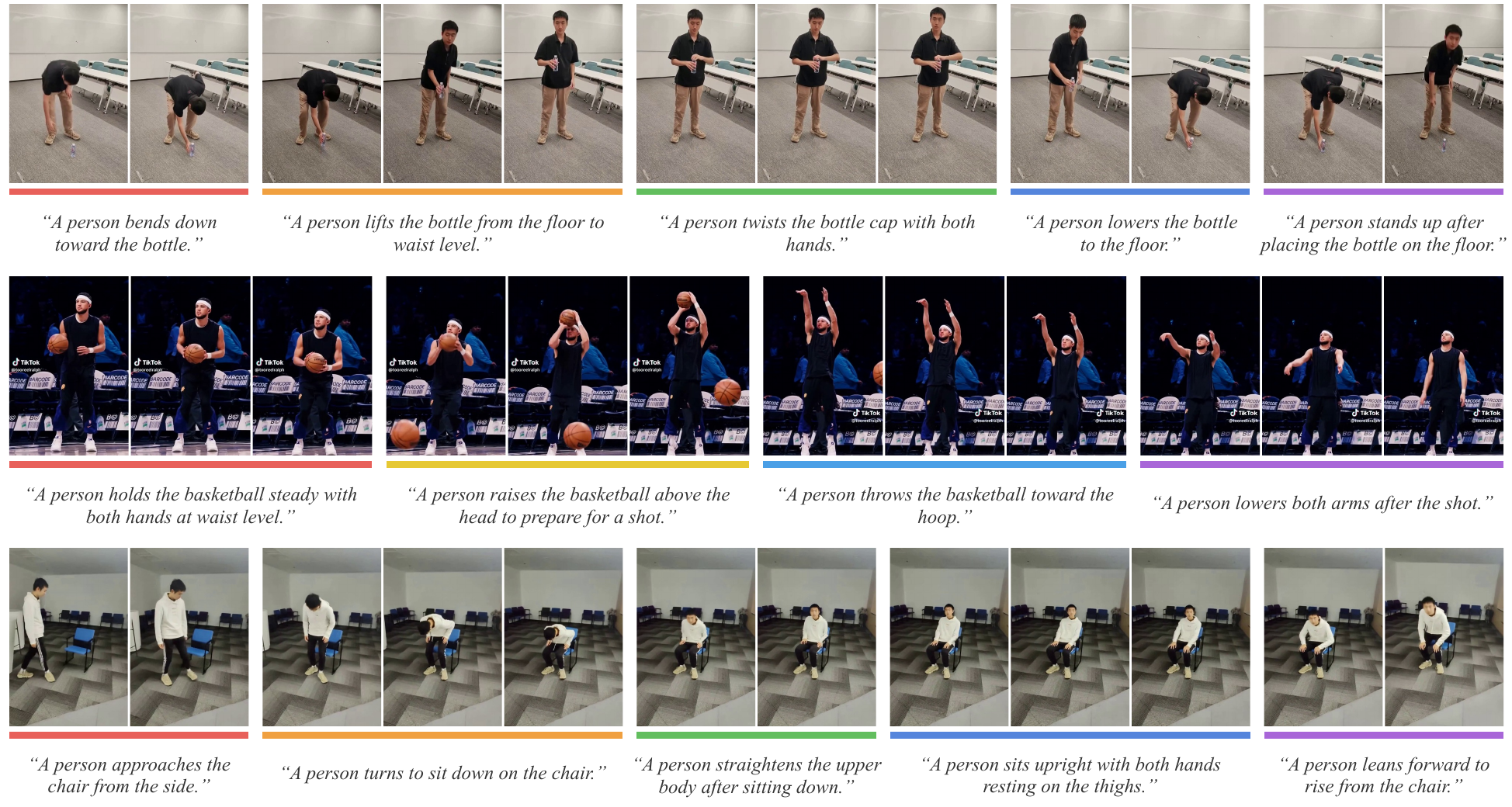}
  \vspace*{-0.6em}
  \caption{\textbf{Qualitative results of video decomposition.}
  Different colors indicate distinct interaction phases.
  }
  \vspace*{-0.4em}
  \label{fig:suppl_decomposition}
\end{figure*}

\noindent\textbf{Text-guided video refinement.}
To refine the rendered novel-view videos, we use Wan2.1-Fun-1.3B-Control~\cite{wan2025wan} as the text-to-video diffusion network.
We first encode the rendered novel-view video into the latent space and inject Gaussian noise with a strength of $0.8$, which alleviates unnatural rendering artifacts while keeping the coarse video structure.
To denoise this noisy video latent, the diffusion network is conditioned on: (1) the semantic prompt $P_{\text{semantic}}$ obtained from motion-aware captioning (\cref{sec:3.2_decomposition}), (2) 2D human skeletons obtained by projecting the initial 3D human motion onto the novel viewpoint, and (3) a reference image that is the first frame of each video segment.
Using these conditions, we iteratively denoise the noisy video latent for 8 steps with a stochastic differential equation (SDE) sampler~\cite{song2021score}.
At each step, the stochastic sampling injects random noise into the video latent and denoises it under the text description.
Through this iterative process, the sampling gradually corrects the implausible human-object interaction in the rendering, based on the prior knowledge of the diffusion network.

\begin{table}[t]
\def\arraystretch{1.6}
\renewcommand{\tabcolsep}{0.6mm}
\footnotesize
\begin{center}
\scalebox{0.72}{
    \begin{tabular}{
    >{\centering\arraybackslash}m{3.5cm}|
    >{\centering\arraybackslash}m{2.6cm}
    >{\centering\arraybackslash}m{2.6cm}
    }
    \specialrule{.1em}{.05em}{0.0em}
        Methods
        & Segment mIoU (\%)${\uparrow}$
        & CLIP similarity${\uparrow}$ \\
        \hline
        w/o fine grounding & 25.78 & 0.741 \\
        w/o motion hint & 43.46 & 0.740 \\
        \textbf{Ours} & \textbf{50.53} & \textbf{0.758} \\
    \specialrule{.1em}{-0.05em}{-0.05em}
    \end{tabular}
}
\end{center}
    \vspace*{-1.5em}
    \caption{\textbf{Evaluation of video decomposition on Charades-STA~\cite{gao2017tall}.}}
    \label{tab:suppl_decomp_eval}
\end{table}

\noindent\textbf{Regional text conditioning.}
To keep the appearance of the human and object faithful to their own descriptions during the refinement, we condition the human and object regions on separate text prompts, $P_{\text{human}}$ and $P_{\text{object}}$.
We derive $P_{\text{human}}$ and $P_{\text{object}}$ from $P_{\text{semantic}}$ using the VLM~\cite{openai2026gpt5}, where $P_{\text{human}}$ describes the identity, clothing, and body shape of the person, and $P_{\text{object}}$ describes the category and shape of the object.
To apply each prompt to its own region, we obtain the human and object masks by rendering the human and object Gaussians separately from each novel viewpoint.
In every cross-attention layer of the diffusion network, we blend the regional prompts into $P_{\text{semantic}}$ with a weight of $\lambda=0.2$, following \cref{eq:regional_text}.
As a result, the refined videos keep the appearance of the human and object faithful to their own descriptions while maintaining their interaction, and serve as reliable supervision for the 4D HOI optimization.

\subsection{4D HOI optimization}
\label{sec:suppl_impl_optimization}
We use the Adam~\cite{kingma2015adam} optimizer for the 4D HOI optimization.
The learning rate is set to $5 \times 10^{-2}$ for the object rotation and translation, $1 \times 10^{-3}$ for the human pose, and $1 \times 10^{-4}$ for the human shape.
We optimize for 250 steps on a single NVIDIA RTX A6000 GPU.

\begin{table}[t]
\def\arraystretch{1.6}
\renewcommand{\tabcolsep}{0.6mm}
\footnotesize
\begin{center}
\scalebox{0.74}{
    \begin{tabular}{
    >{\centering\arraybackslash}m{2.4cm}|
    >{\centering\arraybackslash}m{1.2cm}
    >{\centering\arraybackslash}m{1.2cm}
    >{\centering\arraybackslash}m{1.35cm}
    >{\centering\arraybackslash}m{1.5cm}
    >{\centering\arraybackslash}m{1.3cm}
    >{\centering\arraybackslash}m{1.3cm}
    }
    \specialrule{.1em}{.05em}{0.0em}
        \# of iterations
        & $\text{CD}_{\text{H}}{\downarrow}$
        & $\text{CD}_{\text{O}}{\downarrow}$
        & $\text{Contact}{\uparrow}$
        & $\text{Collision}{\downarrow}$
        & $\text{Accel}_{\text{H}}{\downarrow}$
        & $\text{Accel}_{\text{O}}{\downarrow}$ \\
        \hline
        \textbf{1 (Ours)} & \textbf{5.385} & \textbf{34.894} & \textbf{0.520} & 0.056 & \textbf{0.273} & \textbf{1.571} \\
        2 & 5.419 & 35.392 & 0.515 & 0.056 & 0.277 & 1.590 \\
        3 & 5.675 & 35.652 & 0.512 & \textbf{0.054} & 0.274 & 1.665 \\
    \specialrule{.1em}{-0.05em}{-0.05em}
    \end{tabular}
}
\end{center}
    \vspace*{-1.4em}
    \caption{\textbf{Ablation on iterative re-rendering and optimization.}}
    \label{tab:suppl_iterations}
    \vspace*{-0.6em}
\end{table}
\begin{table}[t]
\def\arraystretch{1.6}
\renewcommand{\tabcolsep}{0.6mm}
\footnotesize
\begin{center}
\scalebox{0.74}{
    \begin{tabular}{
    >{\centering\arraybackslash}m{2.4cm}|
    >{\centering\arraybackslash}m{1.2cm}
    >{\centering\arraybackslash}m{1.2cm}
    >{\centering\arraybackslash}m{1.35cm}
    >{\centering\arraybackslash}m{1.5cm}
    >{\centering\arraybackslash}m{1.3cm}
    >{\centering\arraybackslash}m{1.3cm}
    }
    \specialrule{.1em}{.05em}{0.0em}
        \# of novel views
        & $\text{CD}_{\text{H}}{\downarrow}$
        & $\text{CD}_{\text{O}}{\downarrow}$
        & $\text{Contact}{\uparrow}$
        & $\text{Collision}{\downarrow}$
        & $\text{Accel}_{\text{H}}{\downarrow}$
        & $\text{Accel}_{\text{O}}{\downarrow}$ \\
        \hline
        0 & 5.490 & 64.176 & 0.342 & 0.064 & 0.289 & 3.121  \\
        4 & 5.401 & 39.628 & 0.485 & 0.065 & 0.276 & 1.962 \\
        \textbf{6 (Ours)} & \textbf{5.385} & \textbf{34.894} & \textbf{0.520} & \textbf{0.056} & \textbf{0.273} & \textbf{1.571} \\
        8 & 5.391 & 38.997 & 0.462 & \textbf{0.056} & 0.276 & 1.772 \\
    \specialrule{.1em}{-0.05em}{-0.05em}
    \end{tabular}
}
\end{center}
    \vspace*{-1.4em}
    \caption{\textbf{Ablation on the number of novel views.} }
    \label{tab:suppl_num_views}
\end{table}

\section{Evaluation of video decomposition}
\label{sec:suppl_decomp_eval}
\cref{tab:suppl_decomp_eval} and \cref{fig:suppl_decomposition} demonstrate that our designs in the video decomposition enable accurate localization and text captioning of each interaction phase.
Existing 3D human-object interaction datasets (\textit{e.g.}, Open4DHOI~\cite{wen2025efficient} and BEHAVE~\cite{bhatnagar2022behave}) do not contain text descriptions of each interaction phase.
Instead, we evaluate the video decomposition on 64 videos of Charades-STA~\cite{gao2017tall}, which provides temporal segments of daily human activities annotated with text descriptions.
We measure the accuracy of the interaction phase boundaries with the segment mIoU, which takes the highest temporal IoU with the predicted interaction phases for each GT segment and averages it.
For the text descriptions, we measure the CLIP similarity, which computes the cosine similarity between the CLIP text embeddings~\cite{radford2021learning} of the predicted and GT descriptions at every 1-second interval.
Without the fine boundary grounding, the coarse segments from the object trajectory do not align well with the actual interaction phases, leading to a low segment mIoU.
Without the motion hint in motion-aware captioning, the captions fail to reflect how the object moves relative to the human, resulting in less precise text descriptions and interaction phase boundaries.
By combining both components, our video decomposition achieves the best performance on both metrics.

\section{More ablation studies}
\label{sec:suppl_ablation}
We provide additional ablation studies on the design choices of our framework.
All experiments are conducted on Open4DHOI~\cite{wen2025efficient}, following the same setting as Sec.~\textcolor{cvprblue}{4.3}.

\begin{table}[t]
\def\arraystretch{1.6}
\renewcommand{\tabcolsep}{0.6mm}
\footnotesize
\begin{center}
\scalebox{0.78}{
    \begin{tabular}{
    >{\centering\arraybackslash}m{2.5cm}
    >{\centering\arraybackslash}m{2.5cm}
    >{\centering\arraybackslash}m{2.5cm}
    >{\centering\arraybackslash}m{2.5cm}
    }
    \specialrule{.1em}{.05em}{0.0em}
        VisTracker~\cite{xie2023visibility} 
        & InterTrack~\cite{xie2025intertrack} 
        & CARI4D~\cite{xie2026cari4d} 
        & \textbf{\ourmethod~(Ours)} \\
        \hline
        403 & 783 & 453 & 439 \\
    \specialrule{.1em}{-0.05em}{-0.05em}
    \end{tabular}
}
\end{center}
    \vspace*{-1.4em}
    \caption{\textbf{Running time comparison with video-based methods.} We report the average runtime (sec) per one second of video.}
    \label{tab:suppl_runtime}
\end{table}
\begin{table}[t]
\def\arraystretch{1.6}
\renewcommand{\tabcolsep}{0.6mm}
\footnotesize
\begin{center}
\scalebox{0.72}{
    \begin{tabular}{
    >{\raggedright\arraybackslash}m{2.65cm}|
    >{\centering\arraybackslash}m{1.2cm}
    >{\centering\arraybackslash}m{1.2cm}
    >{\centering\arraybackslash}m{1.35cm}
    >{\centering\arraybackslash}m{1.5cm}
    >{\centering\arraybackslash}m{1.3cm}
    >{\centering\arraybackslash}m{1.3cm}
    }
    \specialrule{.1em}{.05em}{0.0em}
        Methods
        & $\text{CD}_{\text{H}}{\downarrow}$
        & $\text{CD}_{\text{O}}{\downarrow}$
        & $\text{Contact}{\uparrow}$
        & $\text{Collision}{\downarrow}$
        & $\text{Accel}_{\text{H}}{\downarrow}$
        & $\text{Accel}_{\text{O}}{\downarrow}$ \\
        \hline
        TeHOR~\cite{nam2026tehor} & 5.750 & 44.176 & 0.426 & 0.080 & 1.590 & 24.685 \\
        \textbf{\ourmethod~(Ours)} & \textbf{5.385} & \textbf{34.894} & \textbf{0.520} & \textbf{0.056} & \textbf{0.273} & \textbf{1.571} \\
    \specialrule{.1em}{-0.05em}{-0.05em}
    \end{tabular}
}
\end{center}
    \vspace*{-1.4em}
    \caption{\textbf{Comparison with TeHOR~\cite{nam2026tehor} on Open4DHOI~\cite{wen2025efficient}.}}
    \label{tab:suppl_tehor}
\end{table}

\noindent\textbf{Iterative re-rendering and optimization.}
\cref{tab:suppl_iterations} shows that repeating the re-rendering and 4D HOI optimization over multiple iterations does not effectively improve the reconstruction.
Intuitively, since the re-rendering starts from the initial reconstruction, a more accurate reconstruction after the 4D HOI optimization could provide better conditions for the re-rendering.
However, a single pass of re-rendering and optimization is already sufficient to provide reliable supervision for the accurate reconstruction, and additional iterations only increase the running time.
Therefore, considering both performance and computational cost, we choose to perform the re-rendering and optimization only once in our framework.

\noindent\textbf{Number of novel views.}
\cref{tab:suppl_num_views} shows the effect of the number of novel views used in the re-rendering.
In addition to our default setting with six novel views at $\{\pm30^{\circ}, \pm60^{\circ}, \pm90^{\circ}\}$, we compare the settings with zero novel views, four novel views at $\{\pm30^{\circ}, \pm60^{\circ}\}$, and eight novel views at $\{\pm30^{\circ}, \pm60^{\circ}, \pm90^{\circ}, \pm120^{\circ}\}$.
Using more novel views improves the reconstruction, as they provide additional supervision for the regions unobserved in the input video.
However, increasing the number of novel views beyond six rather degrades the reconstruction.
For novel views at large angles, the viewpoint is far from that of the reference image, which leads the re-rendering to produce appearance less consistent with the input video.
Therefore, we use six novel views in our re-rendering pipeline.

\section{Comparison of running time}
\label{sec:suppl_runtime}
\cref{tab:suppl_runtime} shows that \ourmethod~has a competitive running time compared to existing video-based methods.
All methods are evaluated on a single NVIDIA RTX A6000 GPU.
Specifically, the video decomposition, interaction-aware re-rendering, and 4D HOI optimization stages take 10, 308, and 121 seconds, respectively, totaling 439 seconds per one second of video.
Although the re-rendering stage involves a video diffusion network, our framework runs efficiently by requiring only 8 denoising steps and accelerating the subsequent optimization with powerful novel-view cues from the re-rendered videos.

\section{Limitations and future work}
\label{sec:suppl_limitation}

\begin{figure}[t!]
  \centering
  \vspace*{+1.0em}
  \includegraphics[width=0.74\linewidth]{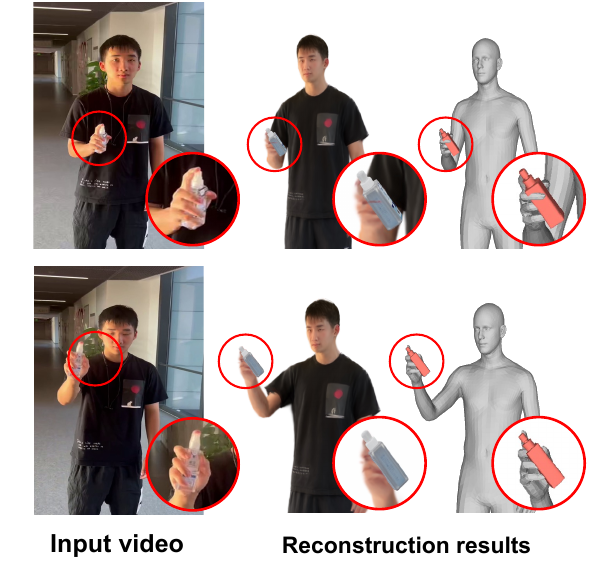}
  \vspace*{-0.34em}
  \caption{\textbf{Failure case of reconstructing local details.}
  }
  \vspace*{-0.4em}
  \label{fig:suppl_limitations}
\end{figure}

\noindent\textbf{Reconstruction of local details.}
While our framework captures the overall interaction between the human and object, it may miss fine-grained local details, such as hand articulation, as shown in \cref{fig:suppl_limitations}.
Since the current video diffusion network~\cite{wan2025wan} struggles to generate complex hand poses, the re-rendered videos provide unreliable cues for hand articulation and its interaction with the object.
A promising future direction is to train an HOI-specific video diffusion network on large-scale human-object interaction videos with diverse hand articulations, providing reliable cues for 4D reconstruction.

\noindent\textbf{Physical grounding.}
Our framework encourages physical plausibility through the contact and collision constraints during the 4D HOI optimization, but does not model physical dynamics, such as gravity and friction.
This limits the use of our reconstructions in robotics applications that require physically valid demonstrations.
Future work could incorporate more physics-based constraints or simulations into the 4D HOI optimization to obtain physically grounded reconstructions.

\section{More comparisons with state-of-the-art methods}
\label{sec:suppl_comparison}
As the image-based methods in \cref{tab:sota_comparison} are evaluated only on key frames to reduce the evaluation cost, we compare \ourmethod~with TeHOR~\cite{nam2026tehor}, a representative image-based method, on every frame of Open4DHOI~\cite{wen2025efficient}.
As shown in \cref{tab:suppl_tehor}, \ourmethod~achieves substantially lower acceleration errors than TeHOR.
While TeHOR reconstructs each frame independently, our framework is supervised by the re-rendered videos, which carry the temporally coherent motion prior of the video diffusion network, yielding temporally consistent 4D reconstructions.

\begin{figure*}[t!]
  \centering
  \includegraphics[width=0.85\linewidth]{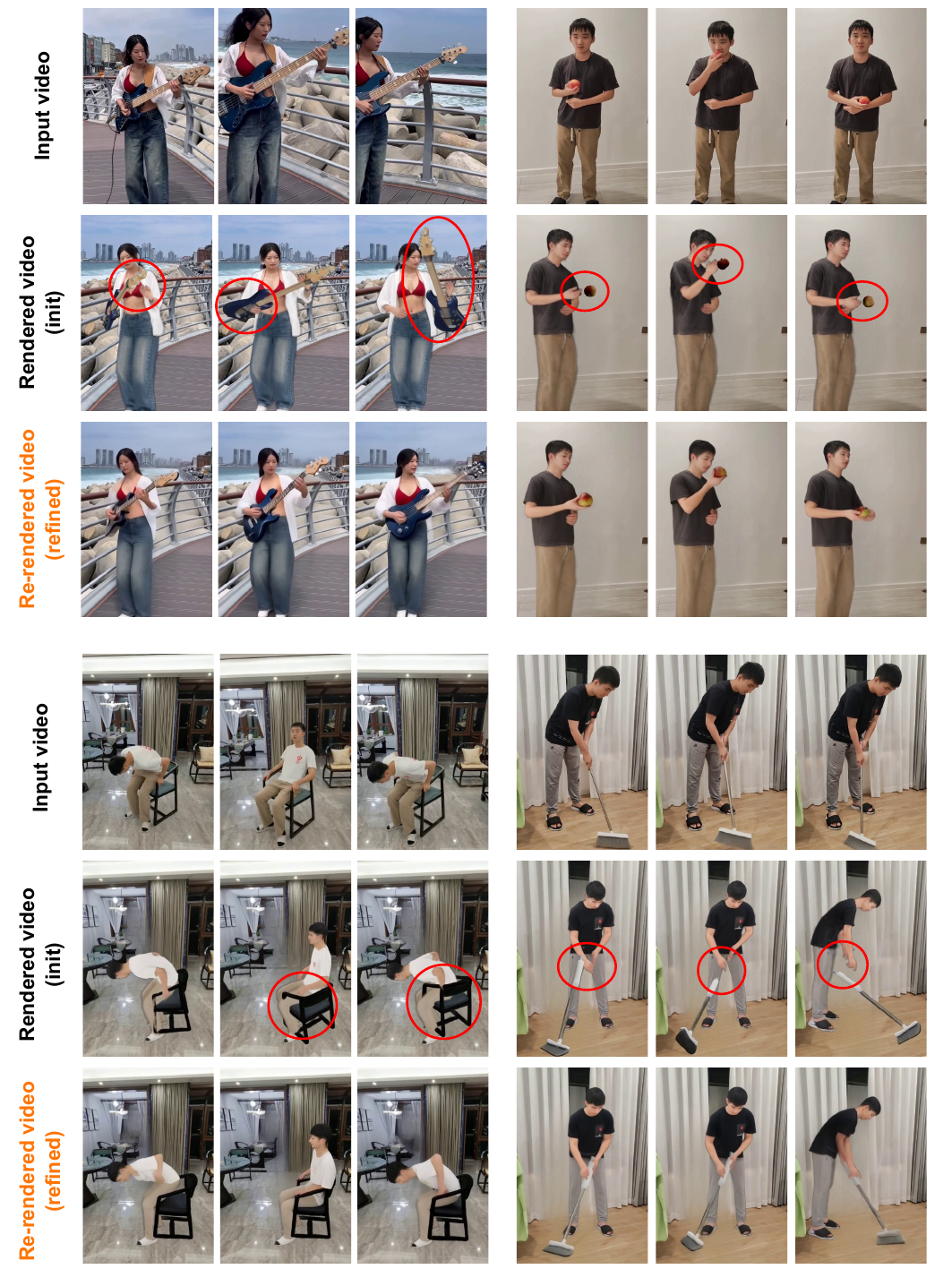}
  \vspace*{-0.0em}
  \caption{\textbf{Qualitative results of re-rendering.}
  We highlight representative failure cases with red circles.
  }
  \label{fig:suppl_rerendering}
\end{figure*}

\begin{figure*}[t!]
  \centering
  \includegraphics[width=0.92\linewidth]{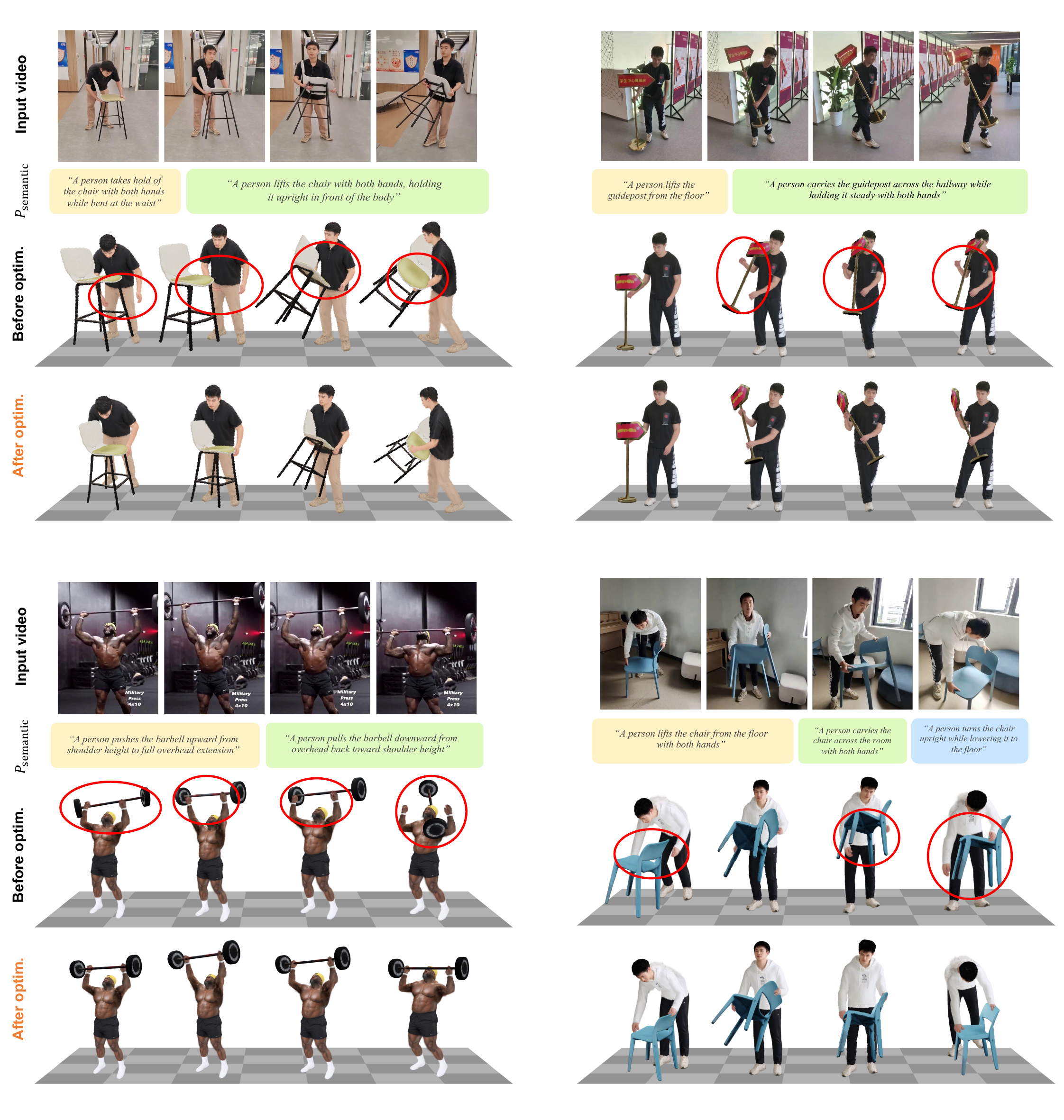}
  \vspace*{-0.8em}
  \caption{\textbf{More qualitative results of 4D HOI optimization.}
  We render all reconstructions at $-30^\circ$ azimuth and highlight representative failure cases with red circles.
  }
  \label{fig:suppl_optimization_results}
\end{figure*}

\begin{figure*}[t]
  \centering
  \includegraphics[width=0.88\linewidth]{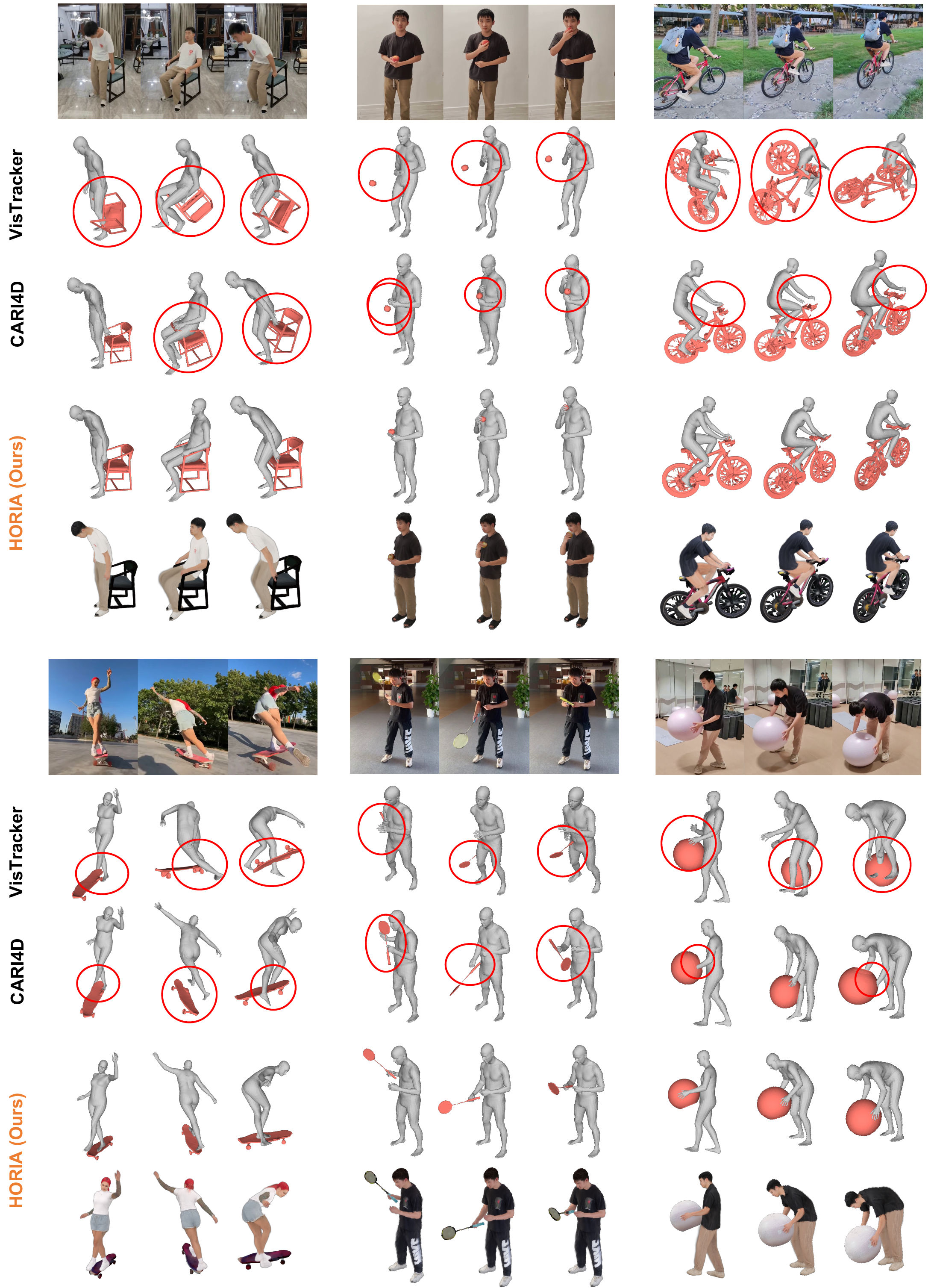}
  \vspace{-0.5em}
  \caption{\textbf{More qualitative comparison with VisTracker~\cite{xie2023visibility} and CARI4D~\cite{xie2026cari4d}.}
  We render each method's reconstruction at $-30^\circ$ azimuth and highlight representative failure cases with red circles.}
  \label{fig:suppl_more_result_1}
\end{figure*}

\begin{figure*}[t]
  \centering
  \includegraphics[width=0.88\linewidth]{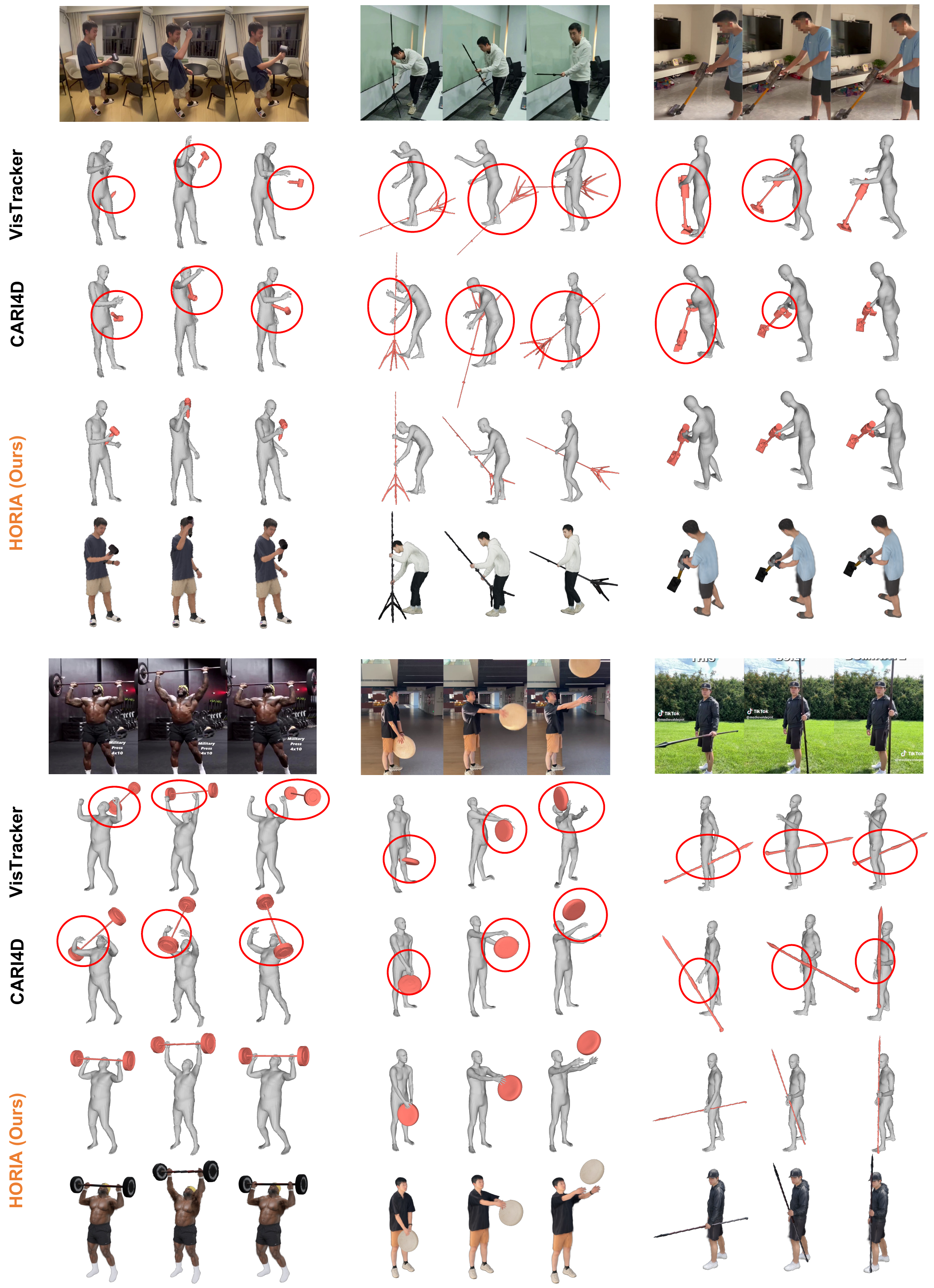}
  \vspace{-0.5em}
  \caption{\textbf{More qualitative comparison with VisTracker~\cite{xie2023visibility} and CARI4D~\cite{xie2026cari4d}.}
  We render each method's reconstruction at $-30^\circ$ azimuth and highlight representative failure cases with red circles.}
  \label{fig:suppl_more_result_2}
\end{figure*}

In addition, \cref{fig:suppl_rerendering} illustrates more additional re-rendering results, which show our method corrects implausible human-object interactions in the initial renderings.
\cref{fig:suppl_optimization_results} presents more results that demonstrate the effectiveness of 4D HOI optimization.
\cref{fig:suppl_more_result_1,fig:suppl_more_result_2} provide more qualitative comparisons with state-of-the-art methods.
These examples demonstrate the effectiveness of our framework in reconstructing the 4D human and object across diverse interaction scenarios.

\clearpage

\clearpage

\end{document}